%% file: main.tex
\documentclass[10pt,twocolumn,letterpaper]{article}
\usepackage{fontawesome5}
\usepackage[pagenumbers]{cvpr}
\usepackage[accsupp]{axessibility}

\input{preamble}

\def\paperID{7677} 
\def\confName{CVPR}
\def\confYear{2026}
 
\title{\method: Feed-Forward 3D Object Articulation}
 
\author{Ruining Li\textsuperscript{1*} \quad  Yuxin Yao\textsuperscript{2*}  \quad   Chuanxia Zheng\textsuperscript{1,3}   \quad   Christian Rupprecht\textsuperscript{1} \\[0.2em]
Joan Lasenby\textsuperscript{2\dag} \quad   Shangzhe Wu\textsuperscript{2\dag}  \quad  Andrea Vedaldi\textsuperscript{1\dag} \\[0.5em]
\textsuperscript{1}University of Oxford  \quad \textsuperscript{2}University of Cambridge  \quad  \textsuperscript{3}Nanyang Technological University \\[0.3em]
{\normalsize \href{https://ruiningli.com/particulate}{\texttt{https://ruiningli.com/particulate}}} 
}

\newif\ifarxiv{}
\arxivtrue{}

\begin{document}

\twocolumn[{%
\renewcommand\twocolumn[1][]{#1}%
\maketitle
\input{figs/teaser}
}]

\def\thefootnote{*}\footnotetext{Equal contribution.
Correspondence to \texttt{ruining@\allowbreak robots.\allowbreak ox.\allowbreak ac.\allowbreak uk} and
\texttt{yy561@\allowbreak cam.\allowbreak ac.\allowbreak uk}.
\quad
\textsuperscript{\dag}{}Equal advising.}\def\thefootnote{\arabic{footnote}}

\input{sec/0_abstract}    
\input{sec/1_intro}
\input{sec/2_related_work}
\input{sec/3_method}
\input{sec/4_experiment}
\input{sec/5_conclusion}

{\small
\bibliographystyle{ieeenat_fullname}
\bibliography{main,vedaldi_general,vedaldi_specific}
}

\clearpage
\input{sec/X_suppl}

\end{document}

%% file: preamble.tex
\usepackage{xspace}
\usepackage{booktabs}
\usepackage{colortbl}
\usepackage{array}
\usepackage{times}
\usepackage{epsfig}
\usepackage{graphicx}
\usepackage{amsmath}
\usepackage{amssymb}
\usepackage{multirow}
\usepackage{adjustbox}
\usepackage{mathtools}
\usepackage[T1]{fontenc}
\usepackage{bookmark}

\definecolor{cvprblue}{rgb}{0.21,0.49,0.74}
\definecolor{Urlcolor}{RGB}{251,111,146}
\definecolor{Linkcolor}{RGB}{193,18,31}
\definecolor{CiteColor}{RGB}{32,126,190}

\definecolor{Urlcolor}{RGB}{251,111,146}
\definecolor{Linkcolor}{RGB}{193,18,31}
\definecolor{CiteColor}{RGB}{32,126,190}

\makeatletter
\renewcommand{\paragraph}{%
\@startsection{paragraph}{4}%
{\z@}{0.2em}{-1em}%
{\normalfont\normalsize\bfseries}%
}
\makeatother

\newcommand{\method}{\textsc{Particulate}\xspace}

\definecolor{ROneColor}{RGB}{37,99,235}     %
\definecolor{RTwoColor}{RGB}{234,88,12}     %
\definecolor{RThreeColor}{RGB}{22,163,74}   %
\definecolor{RFourColor}{RGB}{168,85,247}   %

\newcommand{\best}[1]{\textcolor{Maroon}{\textbf{#1}}}
\newcommand{\secondbest}[1]{\textcolor{RoyalBlue}{\textbf{#1}}}

\providecommand{\faChain}{\faLink}
\providecommand{\faSquareO}{\faSquare[regular]}
\providecommand{\faCut}{\faScissors}
\providecommand{\faLongArrowRight}{\faArrowRight}
\providecommand{\faArrowsV}{\faArrowsAltV}

%% file: figs/teaser.tex
\begin{center}
\vspace{-2em}
\captionsetup{type=figure}
\includegraphics[width=\textwidth]{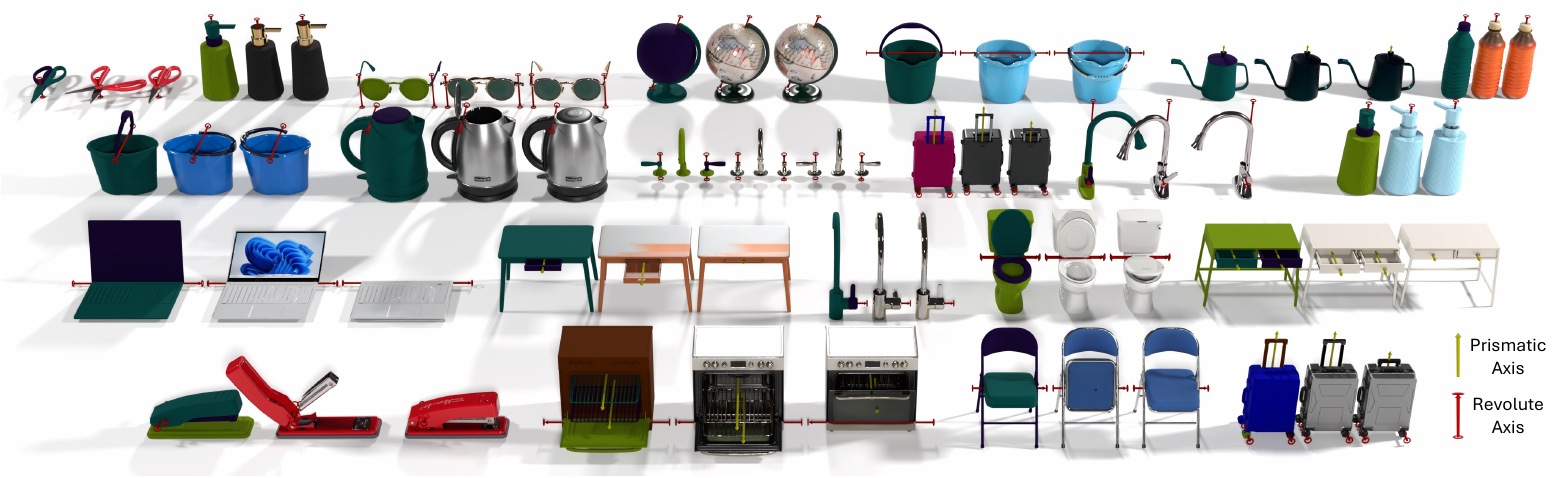}
\captionof{figure}{
\textbf{Articulated 3D objects predicted by \method}.
Our model infers articulated structures directly from static 3D meshes in a single feed-forward pass, enabling fast inference across diverse objects, including those synthesized by 3D generative models.
}%
\label{fig:teaser}
\end{center}

%% file: sec/0_abstract.tex
\begin{abstract}
We introduce \emph{\method}, a feed-forward model that, given a 3D mesh of an object, infers its articulations, including its 3D parts, their kinematic structure, and the motion constraints.
The model is based on a transformer network, the Part Articulation Transformer, which predicts all these parameters for all joints.
We train the network end-to-end on a diverse collection of articulated 3D assets from public datasets.
During inference, \emph{\method} maps the output of the network back to the input mesh, yielding a fully articulated 3D model in seconds, much faster than prior approaches that require per-object optimization.
\emph{\method} also works on AI-generated 3D assets, enabling the generation of articulated 3D objects from a single (real or synthetic) image when combined with an off-the-shelf image-to-3D model.
We further introduce a new challenging benchmark for 3D articulation estimation curated from high-quality public 3D assets, and redesign the evaluation protocol to be more consistent with human preferences.
Empirically, \emph{\method} significantly outperforms state-of-the-art approaches.
\end{abstract}

%% file: sec/1_intro.tex
\section{Introduction}%
\label{sec:intro}

Most objects are defined not only by their shape, but also by their ability to move and deform.
Cabinets, for example, have doors and drawers that open through rotational and translational motion, constrained by hinges and sliding tracks.
Humans can effortlessly understand the \emph{articulated structure} of such objects.
Such capabilities are needed to allow robots to manipulate everyday objects~\cite{xiang20sapien:,li2023behavior,shen2021igibson}, and can simplify the creation of interactive digital twins for gaming and simulation~\cite{jiang2022ditto,weng2024neural}.

In this paper, we consider the problem of estimating the articulated structure of a single static 3D mesh.
Prior work has attempted to model the rich variety of articulated objects via procedural generation~\cite{raistrick2024infinigen, lian2025infinite}.
However, scaling such rule-based approaches to the long tail of real-world objects remains extremely challenging.
We posit that methods that can \emph{learn} articulations from large collections of 3D assets offer greater potential for generality.

Several authors have already proposed learning-based approaches for closely related tasks.
These works~\cite{liu2019rscnn, wang2019dgcnn, liu2023partslip, liu2025partfield} focus on learning part segmentation on 3D point clouds, but primarily predict \emph{semantic} part segmentation on \emph{static} objects, without modeling the articulations between parts.
Other learning-based methods~\cite{lei2023nap,liu2024cage,liusingapo,gao2025meshart} aim to \emph{generate} 3D articulated objects directly.
While promising, these models are typically trained on only a few object categories, assume key attributes such as the kinematic structure are known a priori, and often rely on part retrieval to assemble fully articulated 3D assets, which limits their ability to produce accurate and diverse articulated objects.

Seeking to address these limitations, we introduce \method, a learning-based framework that takes an input 3D mesh and outputs its articulated structure.
\method predicts a full set of articulation attributes, including \emph{articulated} 3D part segmentation, the kinematic structure, and the parameters of the articulated motion.
It does so in a single forward pass, taking only seconds.
We focus on \emph{analyzing} existing 3D assets rather than \emph{synthesizing} them from scratch due to the rapid advances in 3D generative models~\cite{xiang2024structured,deemos24rodin,yang2024hunyuan3d,hunyuan3d22025tencent,lai2025hunyuan3d25highfidelity3d}.
By focusing on the \emph{complementary} task of articulation prediction, \method can leverage increasingly high-fidelity 3D generators to enable one-stop creation of articulated 3D objects with realistic geometry and appearance (\cref{fig:teaser}).
The resulting assets can be seamlessly imported into physics engines for simulation.

\method builds on the Part Articulation Transformer, a flexible and scalable network that takes as input a point cloud approximating the mesh, a representation applicable to virtually any 3D shape.
The network consists of standard attention blocks paired with multiple decoder heads to predict different articulation attributes.
This design allows us to train the network on a diverse collection of articulated 3D assets from public datasets, including PartNet-Mobility~\cite{xiang20sapien:} and GRScenes~\cite{wang2024grutopia}.
To further enhance generalization to novel objects, we augment the raw point cloud with 3D semantic part features obtained from PartField~\cite{liu2025partfield} as input to the network.
The resulting model predicts the articulated structure of new objects in arbitrary articulated poses, including those generated by off-the-shelf 3D generators (\eg,~\cite{lai2025hunyuan3d25highfidelity3d}), in a feed-forward manner.

To evaluate our model, we introduce a new challenging benchmark dataset of $243$ high-quality 3D assets with accurate articulation annotations, crafted by Lightwheel under the CC-BY-NC license~\cite{simready2025}.
We also establish a comprehensive evaluation protocol for the 3D articulation estimation task, with new metrics that more faithfully reflect prediction quality than those used in prior work~\cite{liusingapo,qiu2025articulate}.

In summary, our key contributions are:
(1) \method, a feed-forward model that, given the 3D mesh of an object, infers its full articulated structure, including articulated part segmentation, kinematic structure, and motion parameters, which can then be directly exported to physics simulators.
(2) We show that \method generalizes well to unseen objects, including AI-generated 3D models.
(3) We release a challenging benchmark for articulation estimation.
Experiments show \method's significant gains via metrics that are consistent with human preferences.

%% file: sec/2_related_work.tex
\section{Related Work}%
\label{sec:related}

\input{tables/related_work_supp}

\paragraph{3D part segmentation.}

Recent 3D part segmentation methods increasingly target zero-shot settings~\cite{liu2023partslip,zhou2023partslip++,xue2025zerops,abdelreheem2023satr,tang2024segment}.
Most approaches leverage 2D foundation models such as SAM~\cite{kirillov2023segment} and GLIP~\cite{li2022groundedlanguageimagepretraining} by rendering 3D assets into multiple views and lifting the resulting 2D masks back to 3D.
However, these lift-from-2D approaches, which infer 3D structure from visible surface masks in rendered views, are inherently limited by view coverage and struggle to recover occluded or internal parts.

To alleviate these issues, recent work learns native 3D part segmentation models~\cite{yang2024sampart3d,liu2025partfield,ma2025p3sam,ma20253d}.
These models are trained on large datasets of 3D objects with (pseudo) ground-truth part segmentation, and hence acquire open-world capabilities.
However, they mostly predict \emph{semantic} parts, which often are \emph{not} meaningful for articulation.
By contrast, our \method jointly predicts \emph{articulated} parts with their kinematic structure and motion constraints, producing fully articulated 3D objects from static meshes.

\paragraph{Articulated object modeling.}

Prior work has explored reconstructing articulated objects from multi-view images across different articulation states.
A common paradigm is per-instance optimization using neural radiance fields (NeRFs~\cite{mildenhall20nerf:})~\cite{liu2023paris, wei2022nasam, mu2021sdf, song2024reacto, weng2024neural, wu2022d} or Gaussian splatting (GS~\cite{kerbl233d-gaussian})~\cite{liu2025building, wu2025reartgs}.
These methods capture object-specific articulation well but are slow and require densely sampled, posed images that are difficult to acquire in practice.

To improve scalability, generative models of articulated objects train diffusion~\cite{ho20denoising} or autoregressive models on 3D datasets with annotated articulations~\cite{xiang20sapien:,Geng_2023_CVPR}~\cite{lei2023nap,liu2024cage,liusingapo,Su_2025_CVPR,gao2025meshart}.
However, they are usually trained on only a few categories, assume known attributes such as the kinematic structure, and often rely on part retrieval to assemble articulated assets.

More recent methods leverage foundation models and are training-free or require only light fine-tuning.
DragAPart~\cite{li24dragapart} and DreamArt~\cite{lu2025dreamart} fine-tune pre-trained 2D generators to synthesize articulation videos~\cite{li25puppet-master}; FreeArt3D~\cite{chen2025freeart3d} probes a pre-trained 3D generator~\cite{xiang2024structured}; and Articulate-Anything~\cite{le2024articulate} and Articulate AnyMesh~\cite{qiu2025articulate} prompt vision-language models (VLMs) to reason about part articulation.
Despite their broad generalization across categories, these methods still suffer from slow per-object inference, difficulty with internal or subtle parts, and limited multi-joint support.
Our problem is also related to automatic rigging~\cite{wu23magicpony,li24learning,jakab24farm3d,liu2025riganything,deng2025anymate,song2025magicarticulate}; however, these methods focus primarily on characters, humanoids, animals, or animation-centric assets, and do not directly target the partly-rigid everyday objects we study.

In contrast, we propose a data-driven approach that utilizes a flexible and scalable network to infer all articulation attributes in a single feed-forward pass, enabling recovery of articulated objects in seconds rather than hours and achieving superior performance across diverse categories.

%% file: tables/related_work_supp.tex
\begin{table}[t]
\centering
\small
\setlength{\tabcolsep}{3pt}
\renewcommand{\arraystretch}{1.05}
\resizebox{\linewidth}{!}{%
\begin{tabular}{@{}lcccc|ccccccc|c@{}}
\toprule
& \multicolumn{4}{c|}{\textbf{Input}} & \multicolumn{7}{c|}{\textbf{Output}} & \textbf{Inference} \\
\textbf{Method} & \faLanguage & \faCamera & \faChain & \textbf{3D} & \faSquareO & \textbf{3D} & \faCut & \faChain & \faLongArrowRight & \faArrowsV & $\mathbf{1}^{+}$ & \textbf{time} \\
\midrule

PARIS~\cite{liu2023paris} & &\checkmark& & & & \checkmark & \checkmark &\checkmark &\checkmark & \checkmark & &  N/A\\
NAP~\cite{lei2023nap} & & & & &\checkmark & (\checkmark)* & \checkmark & \checkmark & \checkmark & \checkmark & \checkmark &  N/A\\
MeshArt~\cite{gao2025meshart} & &(\checkmark)$^o$ & & (\checkmark)$^o$& \checkmark & \checkmark & \checkmark &  & \checkmark & \checkmark & \checkmark &  N/A\\
ArtFormer~\cite{Su_2025_CVPR} & \checkmark & \checkmark & & & \checkmark & \checkmark & \checkmark & \checkmark & \checkmark & \checkmark & \checkmark &  N/A \\
ArtiLatent~\cite{chen2025ArtiLatent} & (\checkmark)$^o$ & \checkmark & & & & \checkmark & \checkmark & \checkmark& \checkmark & \checkmark & \checkmark & \textasciitilde 30 sec \\
CAGE~\cite{liu2024cage}& \checkmark & & \checkmark& & \checkmark&  (\checkmark)* & \checkmark & \checkmark & \checkmark&  & \checkmark & N/A \\
SINGAPO~\cite{liusingapo} & & \checkmark & & & \checkmark & (\checkmark)* & \checkmark & \checkmark & \checkmark &\checkmark & \checkmark & \textasciitilde 10 sec \\
\shortstack[c]{Articulate-\\Anything~\cite{le2024articulate}} & (\checkmark)$^o$ &(\checkmark)$^o$ & & & & (\checkmark)*&\checkmark& \checkmark&\checkmark& \checkmark& & \textasciitilde 10 min\\
GEOPARD~\cite{goyal2025geopard}  & & & & \checkmark$^\dagger$ & & \checkmark & & \checkmark & \checkmark & & \checkmark & N/A\\
DreamArt~\cite{lu2025dreamart}  & &\checkmark & & & & \checkmark & \checkmark & & \checkmark & \checkmark & & N/A \\
FreeArt3D~\cite{chen2025freeart3d} &  &\checkmark & \checkmark & & &\checkmark & \checkmark & \checkmark & \checkmark  & & & \textasciitilde 10 min \\
Kinematify~\cite{wang2025kinematify} & (\checkmark)$^o$  & (\checkmark)$^o$ & & & & \checkmark & \checkmark & \checkmark & \checkmark& & \checkmark & \textasciitilde 20 min \\
\shortstack[c]{Articulate\\AnyMesh~\cite{qiu2025articulate}} & & & &\checkmark& &\checkmark&\checkmark&\checkmark&\checkmark& & \checkmark & \textasciitilde 15 min \\
\method (ours) & & & & \checkmark & & \checkmark & \checkmark&\checkmark&\checkmark & \checkmark& \checkmark & \textasciitilde 10 sec \\
\bottomrule
\end{tabular}%
}
\caption{
\textbf{Related work overview on partly-rigid articulated object modeling.}
We summarize existing methods' required inputs and outputs, and their inference time for a single object.
Inputs include
\faLanguage : text,
\faCamera : image or multi-view images,
\faChain : kinematic chain,
\textbf{3D}: 3D geometry (\eg, mesh or point cloud).
Outputs include
\faSquareO : 3D bounding box,
\textbf{3D}: 3D geometry (\eg, mesh or point cloud),
\faCut : part segmentation,
\faChain : kinematic chain,
\faLongArrowRight : motion axis,
\faArrowsV : motion range,
$\mathbf{1}^{+}$: supports multi-joint articulation.
SINGAPO~\cite{liusingapo}, NAP~\cite{lei2023nap}, and CAGE~\cite{liu2024cage} perform part-based retrieval from a 3D part database to obtain articulated 3D objects (\checkmark*).
MeshArt~\cite{gao2025meshart}, ArtiLatent~\cite{chen2025ArtiLatent}, Articulate-Anything~\cite{le2024articulate}, and Kinematify~\cite{wang2025kinematify} all perform unconditional generation and optionally take additional inputs for conditional generation such as texts, images, and 3D point clouds (\checkmark$^o$).
GEOPARD~\cite{goyal2025geopard} further requires segmented 3D point clouds as input (\checkmark$^\dagger$).
N/A indicates the inference time is not reported in the original paper.
}%
\label{tab:related_articulation}
\vspace{-1em}
\end{table}

%% file: sec/3_method.tex
\section{Method}%
\label{sec:method}

\input{figs/method.tex}

We introduce \method, a feed-forward approach that predicts the articulated structure, including the articulated parts, kinematic tree, and motion constraints, from a single static 3D mesh of an object.
We formulate the problem in \cref{sec:method-preliminaries}, followed by details of our network architecture in \cref{sec:method-architecture} and of the decoder heads in \cref{sec:method-decoders}.
We then describe training and inference in \cref{sec:method-training,sec:method-inference}.

\subsection{Problem Definition}%
\label{sec:method-preliminaries}

Given a 3D mesh $\mathcal{M}=(V, F)$ with vertices $V$ and faces $F$, our goal is to predict its (partly rigid) \emph{articulated structure} $\mathcal{A}$.
This specifies
(1) the segmentation $S$ of the mesh faces $F$ into multiple articulated parts;
(2) the kinematic tree $K$ of the parts;
and (3) the rigid motion constraints $M$ of the joints.
Formally, $\mathcal{A}$ is a $4$-tuple $(P, S, K, M)$, where $P\in \mathbb{N}$ is the number of parts and
$
S: [|F|] \to [P]
$
assigns each face to a part, segmenting the mesh $\mathcal{M}$ into $P$ articulated parts\footnote{Note $[n] \coloneq \{1, 2, \dots, n\}$ is the set of positive integers up to $n$.}.
The kinematic tree $K$ is a collection of edges $K\subseteq [P]\times [P]$.
While kinematic trees are undirected by default, for most objects, it is possible and convenient to define a base part (\eg, the casing for a microwave).
We can then orient the edges from the base part outward.
With this, $(p, c) \in K$ indicates a joint connecting part $c$ with its parent part $p$.
We assume a single base part $b$ for each object, and hence $K$ forms an arborescence rooted at $b$.

The motion constraint $M$ is parameterized by 
$
(M_\text{tp},~M_\text{pd},~M_\text{ra},~M_\text{pr},~M_\text{rr})
$,
where each component $M_*$ specifies one aspect of the rigid motion of each part relative to its parent.
Specifically,
$
M_\text{tp}:
[P] \to \{
\operatorname{fixed},
\operatorname{pri},
\operatorname{rev},
\operatorname{both}
\}
$
tells the \textbf{t}y\textbf{p}e of motion:
fixed,
prismatic (allowing linear sliding),
revolute (allowing rotation around a single axis),
or both.
$
M_\text{pd} : [P] \to \mathbb{S}^2
$
gives the direction of each part $p$'s prismatic motion\footnote{
Strictly speaking, $M_\text{pd}$ is only defined on the parts with prismatic motion, \ie,
$
\left \{
p\in[P]:
M_\text{tp}(p)
\in
\left \{
\operatorname{pri},
\operatorname{both}\right \} \right \}
$.
We annotate its domain as $[P]$ for simplicity (similar for $M_\text{ra}$, $M_\text{pr}$ and $M_\text{rr}$ below).
} (\ie, the \textbf{p}rismatic \textbf{d}irections).
$
M_\text{ra} : [P] \to \mathbb{S}^2 \times \mathbb{R}^3
$ 
specifies the rotation axis of each part's revolute motion (\ie, the \textbf{r}evolute \textbf{a}xes).
Both $M_\text{pd}$ and $M_\text{ra}$ are defined in the mesh $\mathcal{M}$'s coordinate system.
$
M_\text{pr} : [P] \to \mathbb{R}^2
$
and
$
M_\text{rr} : [P] \to \mathbb{R}^2
$
define the range $[-l_{\min}, l_{\max}]$ and $[-\theta_{\min}, \theta_{\max} ]$ of each part's  \textbf{p}rismatic \textbf{r}anges and \textbf{r}evolute motion \textbf{r}anges with respect to the mesh $\mathcal{M}$'s articulation state.

The parameterization $(P, S, K, M)$ of $\mathcal{A}$ captures all articulation attributes, and can be easily converted to the Universal Robot Description Format (URDF), allowing \method's predictions to be used in physics simulators.

\subsection{Part Articulation Transformer}%
\label{sec:method-architecture}

Unlike per-shape optimization methods~\cite{qiu2025articulate, lu2025dreamart, wang2025kinematify}, which rely on handcrafted heuristics or distil 2D priors from vision-language foundation models (VLMs), we train a \emph{feed-forward} network $f_\theta$, dubbed Part Articulation Transformer, on a repository of diverse articulated 3D assets to directly predict the articulated structure $\mathcal{A} = f_\theta(\mathcal{M})$ from an input mesh $\mathcal{M}$.
This makes inference fast and better handles small or internal parts that are difficult to view externally and thus difficult to handle by a VLM\@.
Because of these advantages, we can train on \emph{all} articulated objects in existing public datasets.
These span multiple categories and lead to a model more generalizable than prior works~\cite{liusingapo,liu2024cage} that only handled a few similar categories.

To this end, following recent works on 3D deep learning on meshes~\cite{zhang20233dshape2vecset, liu2025partfield, ma2025p3sam}, our model operates on a point cloud $\mathcal{P} = \{\mathbf{p}_i \in \mathbb{R}^3\}_{i=1}^{N}$.
$f_\theta$ processes $\mathcal{P}$, together with a set of learnable part queries, to produce latent point and part vectors using multiple attention blocks.
The latent vectors are later consumed by the decoder heads (\cref{sec:method-decoders}) to predict the articulation attributes $S$, $K$ and $M$.
The overall architecture is illustrated in \cref{fig:method}.

\paragraph{Inputs.}

For each point $\mathbf{p}_i \in \mathcal{P}$, we also provide $f_\theta$ with its associated surface normal $\mathbf{n}_i \in \mathbb{R}^3$ and feature vector $\mathbf{f}_i \in \mathbb{R}^d$ obtained using PartField~\cite{liu2025partfield}.
The normal captures further geometric information, and PartField, which is trained to extract 2D \emph{semantic} parts, captures information.
We use separate MLPs to map these per-point inputs $\mathbf{p}_i$, $\mathbf{n}_i$ and $\mathbf{f}_i$ to vectors of equal dimension $D$ and sum them to obtain each point's token $\tilde{\mathbf{p}}_i \in \mathbb{R}^D$.

\paragraph{Part queries.}

Since the number $P$ of articulated parts in $\mathcal{M}$ is unknown a priori, inspired by~\cite{carion20end-to-end}, we initialize a set of $P_{\max}$ learnable part queries
$
\mathcal{Q} = \{\mathbf{q}_j \in \mathbb{R}^D\}_{j=1}^{P_{\max}}
$,
where $P_{\max}$ is set to be much larger than $P$ in a typical mesh $\mathcal{M}$.

\paragraph{Attention blocks.}

The backbone of $f_\theta$ is a standard transformer~\cite{vaswani17attention}.
Inspired by~\cite{arnaud2025locate}, we apply a sequence of self-attention and cross-attention modules between the point tokens $\{\tilde{\mathbf{p}}_i\}_{i=1}^N$ and the part tokens $\{\mathbf{q}_j\}_{j=1}^{P_{\max}}$.
Specifically, $f_\theta$ consists of $B$ attention blocks alternating
query self-attention and query-to-point cross-attention.
Due to their large number $N \gg P_{\max}$, we do \emph{not} apply self-attention across the point tokens $\{\tilde{\mathbf{p}}_i\}_{i=1}^N$.
This way we can operate on dense point clouds while maintaining a small memory footprint.
The attention blocks output the processed point tokens $\{\tilde{\mathbf{p}}_i\}_{i=1}^N$ and part tokens $\{\tilde{\mathbf{q}}_j\}_{j=1}^{P_{\max}}$.

\subsection{Decoder Heads}%
\label{sec:method-decoders}

We use several prediction heads to map point and part tokens
$\{\tilde{\mathbf{p}}_i\}_{i=1}^N$ and $\{\tilde{\mathbf{q}}_j\}_{j=1}^{P_{\max}}$ to part segmentation, kinematic structure and motion constraints.
Note that the number $P_{\max}$ of part tokens is larger than the number of actual parts $P$.
We will later show how to match the $P_{\max}$ part tokens to the $P$ ground-truth parts for supervision during training (\cref{sec:method-training}) and predict articulated structures $\mathcal{A}$ with $P < P_{\max}$ parts during inference (\cref{sec:method-inference}).
Next, we describe these decoder heads in detail.

\paragraph{Part segmentation.}

$f_\theta$ predicts a matrix of logits $\tilde{\mathbf{S}} \in \mathbb{R}^{N \times P_{\max}}$ telling which points belong to which parts using an MLP $h_{S}$ as
$
\tilde{\mathbf{S}}_{i,j} = h_{S}(\tilde{\mathbf{p}}_i, \tilde{\mathbf{q}}_j)
$.

\paragraph{Kinematic tree.}

The MLP $h_{K}$ takes a pair of part tokens $(\tilde{\mathbf{q}}_i, \tilde{\mathbf{q}}_j)$ and outputs the log-probability that part $i$ is the parent of part $j$, outputting a soft adjacency matrix  $\tilde{\mathbf{K}} \in \mathbb{R}^{P_{\max} \times P_{\max}}$ that captures the kinematic tree:
\begin{equation}
    \tilde{\mathbf{K}}_{i,j}
    \coloneqq \log \mathbb{P}[i \text{ is the parent of } j]
    = h_{K}(\tilde{\mathbf{q}}_i, \tilde{\mathbf{q}}_j).
\end{equation}

\paragraph{Motion types, motion ranges \& prismatic directions.}

Separate MLPs $h_{\text{tp}}$, $h_{\text{pr}}$, $h_{\text{rr}}$ and $h_{\text{pd}}$ take individual part tokens $\tilde{\mathbf{q}}_i$
to predict the motion type, prismatic range, revolute range and prismatic direction of the corresponding part:
\begin{align}
    \tilde{\mathbf{M}}_{\text{tp}}^i &= h_{\text{tp}}(\tilde{\mathbf{q}}_i) \in \mathbb{R}^4, \\
    \tilde{\mathbf{M}}_{\text{pr}}^i &= h_{\text{pr}}(\tilde{\mathbf{q}}_i) \in \mathbb{R}^2, \\
    \tilde{\mathbf{M}}_{\text{rr}}^i &= h_{\text{rr}}(\tilde{\mathbf{q}}_i) \in \mathbb{R}^2, \\
    \tilde{\mathbf{M}}_{\text{pd}}^i &= h_{\text{pd}}(\tilde{\mathbf{q}}_i) / \|h_{\text{pd}}(\tilde{\mathbf{q}}_i)\|_2 \in \mathbb{S}^2.
\end{align}
Here $\tilde{\mathbf{M}}_{\text{tp}}^i \in \mathbb{R}^4$ are the logits of query part $i$ having one of the \emph{four} motion types and $h_\text{pd}(\tilde{\mathbf{q}}_i) \in \mathbb{R}^3$ is normalized to a unit direction vector.

\paragraph{Over-parameterization of revolute axes.}

Next, we consider predicting each part's revolute axis 
$
\tilde{\mathbf{M}}_{\text{ra}}^i \coloneqq (\tilde{\mathbf{d}}_{\text{ra}}^i, \tilde{\mathbf{x}}_{\text{ra}}^i) \in \mathbb{S}^2 \times \mathbb{R}^3
$
where
$\tilde{\mathbf{d}}_{\text{ra}}^i$ denotes the direction and $\tilde{\mathbf{x}}_{\text{ra}}^i$ a fixed point on the axis.
We could use another MLP to do so, but found empirically that this would result in slightly shifted axes.%
\footnote{
Following~\cite{lei2023nap}, we did so by using Pl\"ucker coordinates.
We hypothesize that this is an overfitting problem, as the size of our training data is modest.
The location of the axis $\tilde{\mathbf{x}}_{\text{ra}}^i$ is much more difficult to learn than the direction $\tilde{\mathbf{d}}_{\text{ra}}^i$, probably because the latter is often axis aligned.
}

Because this problem primarily affects location, we do use an MLP $h_{\text{rd}}$ to predict (only) the direction of the revolute axis from the part token $\tilde{\mathbf{q}}_i$ as:
\begin{equation}
    \tilde{\mathbf{d}}_{\text{ra}}^i = h_{\text{rd}}(\tilde{\mathbf{q}}_i) / \|h_{\text{rd}}(\tilde{\mathbf{q}}_i)\|_2.
\end{equation}
However, we \emph{over-parameterize} the location by letting each 3D point $\mathbf{p}_j$ that belongs to the part \emph{vote} for the location of the axis, using an MLP that takes both the point token $\tilde{\mathbf{p}}_j$ and the part token $\tilde{\mathbf{q}}_i$ as input, and maps the point $\mathbf{p}_j$ to its orthogonal projection on the axis:
\begin{equation}
    \tilde{\mathbf{x}}_j^i
    \coloneqq \underset{\mathbf{x}~\text{on the revolute axis}~\tilde{\mathbf{M}}_\text{ra}^i}{\arg\min}
    \|\mathbf{x} - \mathbf{p}_j\|_2
    = h_{\text{cp}}(\tilde{\mathbf{p}}_j, \tilde{\mathbf{q}}_i).
\end{equation}
This leads to a more accurate estimate of
$
\tilde{\mathbf{M}}_{\text{ra}}^i 
= 
(
    \tilde{\mathbf{d}}_{\text{ra}}^i, 
    \tilde{\mathbf{x}}_{\text{ra}}^i
)
$
where  $\tilde{\mathbf{d}}_{\text{ra}}^i$ is predicted directly and $\tilde{\mathbf{x}}_j^i$ is aggregated as explained in \cref{sec:method-inference}.

\subsection{Training}%
\label{sec:method-training}

The network $f_\theta$ is trained end-to-end on public datasets of articulated 3D objects.
For each training object, we sample a point cloud $\mathcal{P}$ and obtain the ground-truth point-to-part segmentation $\mathbf{S}\in \{0,1\}^{N \times P}$, kinematic tree $\mathbf{K}\in \{0,1\}^{P \times P}$ and motion constraints
$
\mathbf{M} = (
    \mathbf{M}_{\text{tp}} \in \{0,1\}^{P \times 4},
    \mathbf{M}_{\text{pd}} \in \mathbb{R}^{P \times 3},
    \mathbf{M}_{\text{pr}} \in \mathbb{R}^{P \times 2},
    \mathbf{M}_{\text{rr}} \in \mathbb{R}^{P \times 2},
    \mathbf{d}_{\text{ra}} \in \mathbb{R}^{P\times 3},
    \mathbf{x} \in \mathbb{R}^{N\times 3}
)
$
to supervise $f_\theta$.

\paragraph{Matching predicted parts with ground-truth.} 

For training, we need to map each ground-truth part $i \in [P]$ to a distinct part query $j = \pi(i) \in [P_{\max}]$.
Following DETR~\cite{carion20end-to-end}, the assignment $\hat{\pi} : [P] \to [P_{\max}]$ (an injection) is inferred by aligning the ground-truth and predicted point-to-part assignments $\mathbf{S}_{j,i}$ and $\tilde{\mathbf{S}}_{j,k}$ as
\begin{equation}
\hat{\pi} = \arg\max_{\pi} \sum_{i=1}^P
\sum_{j=1}^N \mathbf{S}_{j,i} \log \frac{\exp(\tilde{\mathbf{S}}_{j,\pi(i)})}{\sum_{k=1}^{P_{\max}} \exp(\tilde{\mathbf{S}}_{j,k})}.
\end{equation}
The assignment $\hat{\pi}$ is computed on the fly using the Hungarian algorithm and used to map the ground truth articulation attributes of each part to the corresponding query for supervision, as explained next.

\paragraph{Training losses.}

We train the network $f_\theta$ end-to-end using a multi-task loss:
$
\mathcal{L}=\mathcal{L}_S + \mathcal{L}_K + \mathcal{L}_{M},
$
where
$
\mathcal{L}_{M} =
\mathcal{L}_{M_\text{tp}} +
\mathcal{L}_{M_\text{pr}} +
0.1 \mathcal{L}_{M_\text{rr}} +
\mathcal{L}_{M_\text{pd}} +
\mathcal{L}_{d_\text{ra}} +
\mathcal{L}_{x_\text{ra}}
$
is the combined loss for motion constraints.
$
\mathcal{L}_S = \operatorname{CE}(\tilde{\mathbf{S}}^*, \mathbf{S})
$
is the cross-entropy loss for the part segmentation, where $\tilde{\mathbf{S}}^* \in \mathbb{R}^{N \times P}$ is obtained by column-wise permuting $\tilde{\mathbf{S}}$ according to $\hat{\pi}$
(\ie, $\tilde{\mathbf{S}}^*_{j,i} = \tilde{\mathbf{S}}_{j,\hat{\pi}(i)}$, and analogously for the starred notations below).
$\mathcal{L}_K = \operatorname{BCE}(\tilde{\mathbf{K}}^*, \mathbf{K})$ is the binary cross-entropy loss for the kinematic tree.
The terms in $\mathcal{L}_M$ supervise the motion parameters, comparing each part query's prediction with its matched ground truth:
$
\mathcal{L}_{M_\text{tp}} = \operatorname{CE}(\tilde{\mathbf{M}}_{\text{tp}}^*, \mathbf{M}_{\text{tp}})
$,
$
\mathcal{L}_{M_\text{pr}} = \|\tilde{\mathbf{M}}_{\text{pr}}^* - \mathbf{M}_{\text{pr}}\|_1
$,
$
\mathcal{L}_{M_\text{rr}} = \|\tilde{\mathbf{M}}_{\text{rr}}^* - \mathbf{M}_{\text{rr}}\|_1
$,
$
\mathcal{L}_{M_\text{pd}} = \|\tilde{\mathbf{M}}_{\text{pd}}^* - \mathbf{M}_{\text{pd}}\|_1
$,
$
\mathcal{L}_{d_\text{ra}} = \|\tilde{\mathbf{d}}_{\text{ra}}^* - \mathbf{d}_{\text{ra}}\|_1
$,
and
$
\mathcal{L}_{x_\text{ra}} = \|\tilde{\mathbf{x}} - \mathbf{x}\|_1
$.

\subsection{Inference}%
\label{sec:method-inference}

We now describe how \method infers the 3D articulated structure $\mathcal{A}$ from a static mesh $\mathcal{M}$.

\paragraph{Feed-forward prediction.}

Given the input mesh $\mathcal{M} = (V, F)$, we first sample a point cloud $\mathcal{P}$, ensuring that each face has at least one point $\mathbf{p}_i \in \mathcal{P}$ sampled on it.
We then feed $\mathcal{P}$ into $f_\theta$ to map each point $\mathbf{p}_i$ to a corresponding part index $\tilde{\mathbf{s}}_i = \arg\max_{j\in [P_{\max}]} \tilde{\mathbf{S}}_{i,j}$.
Similarly, we assign each face $f \in [|F|]$ to the part index $S(f)$ of its majority points (breaking ties randomly).

Note that the number of actual parts $P$ in $\mathcal{A}$ is often much smaller than $P_{\max}$.
The set of active part indices is simply the \emph{range}
$
\mathcal{Q}_{\mathcal{A}} = \{S(f) :  f \in [|F|] \} \subseteq [P_{\max}]
$
of the face assignments $S$, and the number of active parts is $P = |\mathcal{Q}_{\mathcal{A}}|$.
Next, we compute the kinematic tree $K$ from $f_\theta$'s prediction $\tilde{\mathbf{K}} \in \mathbb{R}^{P_{\max} \times P_{\max}}$, where $\tilde{\mathbf{K}}_{i,j}$ represents the log likelihood of part query $i$ being the parent of part query $j$.
To do so, we run Edmonds' algorithm to extract the \emph{maximum} spanning arborescence (MSA) of $\tilde{\mathbf{K}}$.
The induced arborescence of this MSA on the present parts $\mathcal{Q}_{\mathcal{A}}$ forms $K$.

We can directly read off the network predictions 
$
\tilde{\mathbf{M}}_{\text{tp}}^{\mathcal{Q}_{\mathcal{A}}}
$, 
$\tilde{\mathbf{M}}_{\text{pd}}^{\mathcal{Q}_{\mathcal{A}}}$,
$\tilde{\mathbf{M}}_{\text{pr}}^{\mathcal{Q}_{\mathcal{A}}}$ and
$\tilde{\mathbf{M}}_{\text{rr}}^{\mathcal{Q}_{\mathcal{A}}}$ to get the motion types $M_\text{tp}$,
prismatic directions $M_\text{pd}$,
prismatic ranges $M_\text{pr}$ and 
revolute ranges $M_\text{rr}$ of the present parts $\mathcal{Q}_{\mathcal{A}}$.
For the revolute axes, we first infer their directions as $\tilde{\mathbf{d}}_{\text{ra}}^{\mathcal{Q}_\mathcal{A}}$.
Then, for each point $\mathbf{p}_j \in \mathcal{P}$, we use the MLP $h_{\text{cp}}$ to predict the closest point on the axis of its part $\tilde{\mathbf{s}}_j$, so that each point effectively \emph{votes} for the axis location.
We take the (coordinate-wise) \emph{median} of these votes as the axis location
Formally, $M_\text{ra}(i)  \in \mathbb{S}^2 \times \mathbb{R}^3$ is computed as:
\begin{equation}
    M_\text{ra}(i) = \left(
        \tilde{\mathbf{d}}_{\operatorname{ra}}^i,
        \operatorname{median}(\{h_\text{cp}(\tilde{\mathbf{p}}_j, \tilde{\mathbf{q}}_i) : \tilde{\mathbf{s}}_j = i\})
    \right)
\end{equation}
for $i \in \mathcal{Q}_{\mathcal{A}}$.
Together, the inferred tuple $(S, K, M)$ where $M = (M_{\text{tp}}, M_{\text{pd}}, M_{\text{ra}}, M_{\text{pr}}, M_{\text{rr}})$ defines the 3D articulated structure $\mathcal{A}$.

\paragraph{Refinement with connected components.}

Many handcrafted 3D assets have well-defined connected components, which we can leverage to refine the segmentation.
For such input meshes, after obtaining the preliminary per-face part segmentation $S$, we ensure that all faces within the same connected component $\mathcal{C}$ are assigned to the same part, selected as the one with the largest surface coverage in $\mathcal{C}$.

%% file: figs/method.tex
\begin{figure*}[t]
\centering
\includegraphics[width=\textwidth]{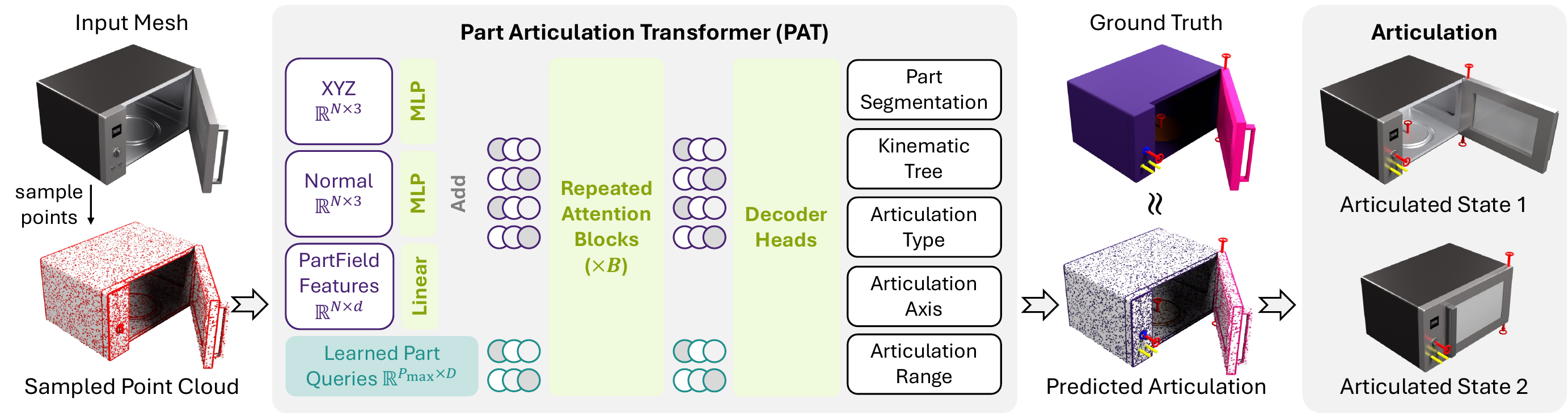}
\caption{
\textbf{Overview of \method.}
Our model consumes a point cloud sampled from the input mesh and predicts its articulated structure.
The backbone consists of $B$ standard attention blocks that operate on the point vectors and a set of learnable part query vectors.
The resulting latent vectors are then processed by specialized decoder heads dedicated to different components of the articulated structure. 
The entire model is trained end-to-end in a fully supervised manner.
At inference time, \method projects the point-level segmentation predictions back onto the original mesh to reconstruct a complete articulated 3D model.
\label{fig:method}
}%
\vspace{-1em}
\end{figure*}

%% file: sec/4_experiment.tex
\section{Experiments}%
\label{sec:experiments}

We evaluate \method on recovering articulated structures from static 3D meshes and compare it with prior work that addresses this task in full or in part (\cref{sec:experiments-results}).
We structure the evaluation around two sub-tasks: articulated part segmentation (\cref{sec:compare_segment}) and articulated motion prediction (\cref{sec:compare_articulation}), so we can compare our all-in-one approach with methods that handle only one of them.
In~\cref{sec:experiments-ablation}, we discuss the contribution of data and network design choices.

\subsection{Experiment Details}%
\label{sec:experiments-details}

\paragraph{Training datasets.}

\method is trained on the PartNet-Mobility~\cite{xiang20sapien:} and GRScenes~\cite{wang2024grutopia} datasets.
For~\cite{xiang20sapien:}, we follow the train-test split by~\cite{liusingapo}.
For~\cite{wang2024grutopia}, we use all its articulated assets.
We set $P_{\max}{=}16$ and discard objects with more than $16$ articulated parts.
This yields $3,800$ objects across $50$ categories.

\paragraph{Evaluation datasets.}

We evaluate \method on the test split of PartNet-Mobility~\cite{xiang20sapien:}, which contains $7$ common categories.
In addition, we introduce a new challenging benchmark for the 3D articulation estimation task, which contains $243$ high-quality articulated objects spanning $14$ categories crafted by Lightwheel and released under a CC-BY-NC license~\cite{simready2025}.

\paragraph{Implementation details.}

During training, our model is given point clouds of a fixed size $N{=}2048$, among which $50\%$ are sampled uniformly on the mesh surface and the remaining $50\%$ are sampled from the \emph{sharp edges} where dihedral angles are larger than $30^\circ$~\cite{chen2025dora}.
We sample a random articulated state for each training iteration and compute the PartField~\cite{liu2025partfield} features on the fly.
All point clouds are first normalized to $[-0.5, 0.5]^3$, followed by data augmentation.
Each point cloud is scaled by a factor randomly drawn from $\mathcal{U}(0.95, 1.05)$ and translated by a vector from $\mathcal{N}(0, 0.02)^3$.
Our model consists of $B{=}8$ attention blocks with a total of $150$M parameters.
The final model is optimized for $100$K iterations using AdamW with a global batch size of $128$ on $8$ H100 GPUs.
At inference time, the model receives point clouds of size $102,400$ sampled from the input mesh using the same procedure to ensure coverage of all mesh faces.

\subsection{Results and Comparisons}%
\label{sec:experiments-results}

\input{figs/fig-qualitative-results}

\subsubsection{Qualitative Results on Synthesized 3D Meshes}

In~\cref{fig:qualitative-results}, we visualize the articulated 3D objects of a variety of categories predicted by \method from static 3D meshes generated by Hunyuan3D~\cite{hunyuan3d25hunyuan3d}, an off-the-shelf 3D generator.
Our model recovers articulated parts faithful to the kinematic structure of the input objects, with plausible motion constraints for the movable parts.
Notably, despite being trained only on artist-created assets, our model generalizes well to AI-generated meshes, yielding a fully automatic pipeline for generating articulated 3D assets directly from text prompts or images.

\subsubsection{Comparisons on Articulated Part Segmentation}%
\label{sec:compare_segment}

\paragraph{Baselines.}

We compare against four state-of-the-art 3D part segmentation methods:
\textbf{PartField}~\cite{liu2025partfield}, \textbf{P3SAM}~\cite{ma2025p3sam}, \textbf{SINGAPO}~\cite{liusingapo}, and \textbf{Articulate AnyMesh}~\cite{qiu2025articulate}.
For reference, we also include a \textbf{Naive Baseline}, which simply treats the entire object as a single (fixed) part.
As PartField, P3SAM, and Articulate AnyMesh leverage face connectivity for part segmentation, we likewise refine SINGAPO and our outputs using mesh connected components as described in~\cref{sec:method-inference}, and report metrics both with and without this refinement.
PartField allows specifying the number of output parts, so we provide it with the number of ground-truth (GT) parts.
P3SAM additionally requires point prompts, so we supply one point randomly sampled from each GT part.
SINGAPO does \emph{not} take a 3D mesh as input, and instead generates bounding boxes for the parts conditioned on a single object-centric image and retrieves their geometries from a repertoire of part meshes.
To reduce any disadvantage from not having access to the original mesh, we make the following adjustments when evaluating SINGAPO\@:
first, instead of one-shot generation, we generate $10$ samples per instance using different rendered views for conditioning, and report a ``1@10'' metric computed using the best sample among the $10$;
second, on the Lightwheel evaluation set, we extract part geometries directly as sub-meshes of the original mesh within the generated bounding boxes, rather than retrieving them from the part repository curated from the PartNet-Mobility training set.
We use the official implementations of all baselines and the model weights publicly released by the authors without further finetuning.

\paragraph{Metrics.}%
\label{sec:metrics}

We report three metrics for part segmentation: generalized Intersection over Union (\textbf{gIoU})~\cite{rezatofighi2019generalized}, mean Intersection over Union (\textbf{mIoU}), and bidirectional part-wise Chamfer distance (\textbf{PC}), computed between predicted and GT parts.
Since the predicted and GT parts have unknown correspondence and may differ in count, we first perform Hungarian matching between the two sets based on pairwise part-centroid distances.
Prior work~\cite{liusingapo,liu2024cage,lei2023nap} computes metrics only on matched part pairs and ignores unmatched parts.
However, as we show in~\cref{sec:appendix_results_eval_details}, metrics computed in this manner do \emph{not} sufficiently penalize missing small parts:
under this protocol, the Naive Baseline outperforms \emph{all} baseline methods on \emph{all} metrics, making these scores uninformative for assessing part-segmentation quality.
To better assess the full set of predictions, we introduce a penalty for any predicted or GT part that remains unmatched.
Specifically, we compute each metric $\mathcal{D}_d$ over all predicted parts $\{p_i^{\text{pred}}\}_{i=1}^{N}$ and GT parts $\{p_j^{\text{gt}}\}_{j=1}^{M}$:
\begin{equation}
   \mathcal{D}_d =  \frac{1}{2}\left(\frac{1}{N} \sum_{i=1}^{N}\tilde{d}(p_i^{\text{pred}}) + \frac{1}{M} \sum_{j=1}^{M}\tilde{d}(p_j^{\text{gt}})\right),
\end{equation}
where $\tilde{d}(p_i)=d\!\left(p_i, \mathcal{H}(p_i)\right)$ if (predicted or GT) part $p_i$ is matched to a (GT or predicted) part $\mathcal{H}(p_i)$, and is otherwise set to a penalty value $\epsilon{=}-1$ for gIoU, $0$ for mIoU, and half the diagonal length of the input mesh's bounding box for PC\@.
Here, $d$ selects the metric (gIoU, mIoU, or PC), and $\mathcal{D}_d$ reports that metric with penalties applied to unmatched parts.
We report $\mathcal{D}_d$ averaged over all instances in the test set as the final score, where $d$ specifies the metric (gIoU, mIoU, or PC).
All metrics are evaluated based on $10^6$ points uniformly sampled from the input mesh.

\input{tables/main_segmentation}
\paragraph{Results.}
We report the quantitative results on the PartNet-Mobility test set and our Lightwheel benchmark in \cref{tab:rest-comparison}, where \method consistently outperforms all baselines on both datasets.
Note that PartField and P3SAM are given privileged information at test time including the exact number of GT parts, therefore avoiding any penalty for unmatched parts.
All methods except P3SAM experience a performance drop on the more challenging Lightwheel dataset, which contains more diverse assets with substantially finer part annotations.

\input{figs/fig-qualitative-comp}

In \cref{fig:qualitative-comp}, we visualize the part segmentation results of \method and the baselines on the Lightwheel benchmark.
PartField and P3SAM are trained for \emph{semantic} part segmentation, where part definitions differ from those of \emph{articulated} parts.
Consequently, their predicted segments often fail to coincide with the GT parts (\cref{fig:qualitative-comp}b).
SINGAPO is trained only on a limited number of categories, and hence fails to generalize to more diverse objects (\cref{fig:qualitative-comp}c).
Articulate AnyMesh's VLM-based pipeline cannot handle internal parts which are invisible under the rest state (\cref{fig:qualitative-comp}a and \cref{fig:qualitative-comp}b).
By contrast, our data-driven approach accurately recovers articulated parts, including internal ones, across a wide variety of object categories.

\input{tables/main_articulation}

\subsubsection{Comparisons on Articulated Motion Prediction}%
\label{sec:compare_articulation}

We now compare our method's predicted motion constraints (\ie, types, axes and ranges) with those of the baselines.

\paragraph{Baselines.}

Out of the four baselines we compare for part segmentation, \textbf{SINGAPO}~\cite{liusingapo} and \textbf{Articulate AnyMesh}~\cite{qiu2025articulate} also predict articulated motion, which we compare here.
Note that Articulate AnyMesh predicts the motion type and axis for each part but does not estimate its range.
For quantitative evaluation, we assign each predicted part the motion range of its matched GT part, using the same Hungarian matching based on part-centroid distances.

\paragraph{Metrics.}

Since the predicted parts may differ substantially from the GT parts, directly comparing the predicted motion parameters with those of the matched GT parts is not always sensible.
Following prior work~\cite{liu2024cage, liusingapo}, we instead compare the resulting articulated geometries.
Specifically, given the predicted articulated parts and their motion constraints, we fully articulate the predicted articulated asset, where every part is moved to its maximum extent.
We then compare the resulting fully articulated geometry with the ground-truth counterpart also in its fully articulated state, and report the part-wise \textbf{gIoU} and Chamfer distance (\textbf{PC}) from \cref{sec:compare_segment} as proxies for motion accuracy.
As in the segmentation metrics, unmatched parts incur a penalty.
Furthermore, we introduce a new metric, \emph{whole-object} Chamfer distance (\textbf{OC}), which measures the overall bidirectional Chamfer distance between the entire predicted geometry and its GT counterpart in the fully articulated state.

\paragraph{Results.}
The results are reported in \cref{tab:full-comparison}, where \method significantly outperforms all baselines across all metrics on both datasets.
This includes Articulate AnyMesh, which is given the ground-truth motion range for each predicted part.
These results demonstrate the superiority of our end-to-end approach in estimating articulated motion constraints.

\input{tables/ablation_lightwheel}

\subsection{Ablations}%
\label{sec:experiments-ablation}

\paragraph{Data.}

We train a separate model, using identical hyperparameters, on the PartNet-Mobility~\cite{xiang20sapien:} dataset \emph{only}.
In \cref{tab:ablation} ($\mathbb{A}$ \vs $\mathbb{B}$), reducing the training set leads to a slightly worse model.
The model still outperforms the baselines by a wide margin, indicating that ours gains mainly come from the simple, data-efficient design.

\paragraph{Over-parameterization of revolute axes.}

We evaluate the effectiveness of the proposed over-parameterization of revolute axes by comparing it with an alternative design 
in which an MLP directly regresses each axis in $\mathbb{R}^6$ following~\cite{lei2023nap}.
In \cref{tab:ablation} ($\mathbb{A}$ \vs $\mathbb{C}$), over-parameterization improves performance, as it mitigates overfitting and averages out errors in individual location estimates.

\paragraph{Pre-trained semantic features.}

In~\cref{tab:ablation} ($\mathbb{A}$ \vs $\mathbb{D}$), PartField features improve performance, as they provide global shape context and richer semantic cues.

%% file: figs/fig-qualitative-results.tex
\begin{figure*}[t]
\centering
\includegraphics[trim={40pt 0 40pt 0}, clip, width=\textwidth]{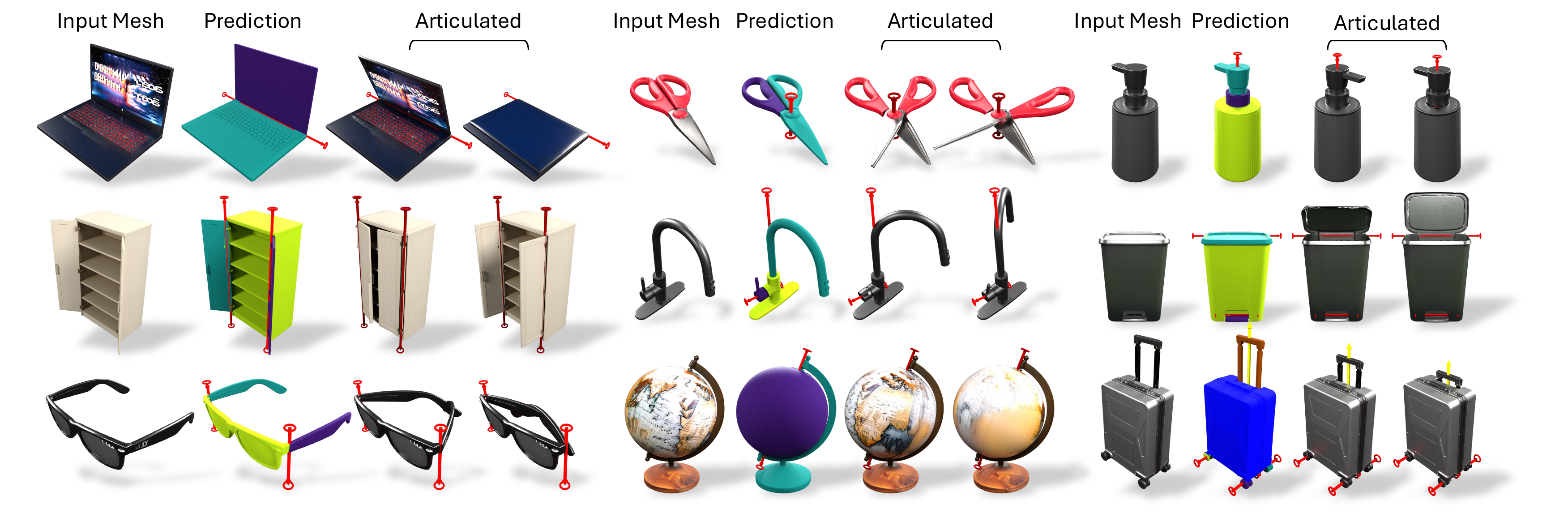}
\vspace{-2em}
\caption{
\textbf{Qualitative results} of \method, illustrating the input meshes and the predicted articulated parts with their motion constraints.
We also show each 3D object in two different articulated states.
All input meshes are \emph{generated} using an off-the-shelf 3D generator.
}%
\label{fig:qualitative-results}
\vspace{-1em}
\end{figure*}

%% file: tables/main_segmentation.tex
\begin{table}[t]
\centering
\small
\setlength{\tabcolsep}{3pt}
\renewcommand{\arraystretch}{1.05}
\resizebox{\linewidth}{!}{%
\begin{tabular}{@{}lcccccc@{}}
\toprule
 & \multicolumn{3}{c}{\textbf{Lightwheel}} & \multicolumn{3}{c}{\textbf{PartNet-Mobility}} \\
\cmidrule(lr){2-4}\cmidrule(lr){5-7}
\textbf{Method}
& gIoU$\uparrow$
& PC$\downarrow$
& mIoU$\uparrow$
& gIoU$\uparrow$
& PC$\downarrow$
& mIoU$\uparrow$ \\
\midrule
Naive Baseline & 0.018 & 0.285 & 0.413 & 0.296 & 0.210 & 0.612 \\
SINGAPO~\cite{liusingapo}        & -0.116 & 0.201 & 0.272 & -- & -- & -- \\
SINGAPO~\cite{liusingapo} (1@10)        & -0.096 & 0.190 & 0.277 & -- & -- & -- \\
\method (ours)           & \secondbest{0.183} & \best{0.163} & 0.430 & \secondbest{0.879} & \secondbest{0.003} & \secondbest{0.883} \\
\midrule
PartField~\cite{liu2025partfield}$^\dagger$  & 0.079 & 0.106 & 0.264  & 0.183 & 0.123 & 0.361 \\
P3SAM~\cite{ma2025p3sam}$^\dagger$     & 0.122 & 0.177 & 0.411 & -0.116 & 0.261 & 0.267 \\
SINGAPO~\cite{liusingapo}$^\dagger$   & -0.097 & 0.234 & 0.273 & 0.262 & 0.124 & 0.468 \\
SINGAPO~\cite{liusingapo} (1@10)$^\dagger$  & -0.050 & 0.221 & 0.297 &   0.271 & 0.117 & 0.471 \\
Articulate AnyMesh~\cite{qiu2025articulate}$^\dagger$   & 0.172 & 0.190 & \secondbest{0.452} & 0.383 & 0.104 & 0.542 \\
\method (ours)$^\dagger$ & \best{0.332} & \secondbest{0.168} & \best{0.576} & \best{0.880} & \secondbest{0.003} & \best{0.884}\\
\bottomrule
\end{tabular}%
} %
\caption{
\textbf{Part segmentation results}.
We report generalized IoU (gIoU), part-wise Chamfer distance (PC), and mean IoU (mIoU), all computed with penalties applied to unmatched parts.
$^\dagger$: leveraging mesh connectivity.
--: SINGAPO retrieves part geometries from a part library on the PartNet-Mobility test set, which assumes mesh connectivity. Therefore, its metrics without leveraging connectivity are left empty.
1@10: computed using the best sample out of 10 predictions per instance.
Colors: \best{best} and \secondbest{second best}.
}%
\label{tab:rest-comparison}
\vspace{-1em}
\end{table}

%% file: figs/fig-qualitative-comp.tex
\begin{figure}[t]
\centering
\includegraphics[trim={0pt 0 50pt 0}, clip, width=\columnwidth]{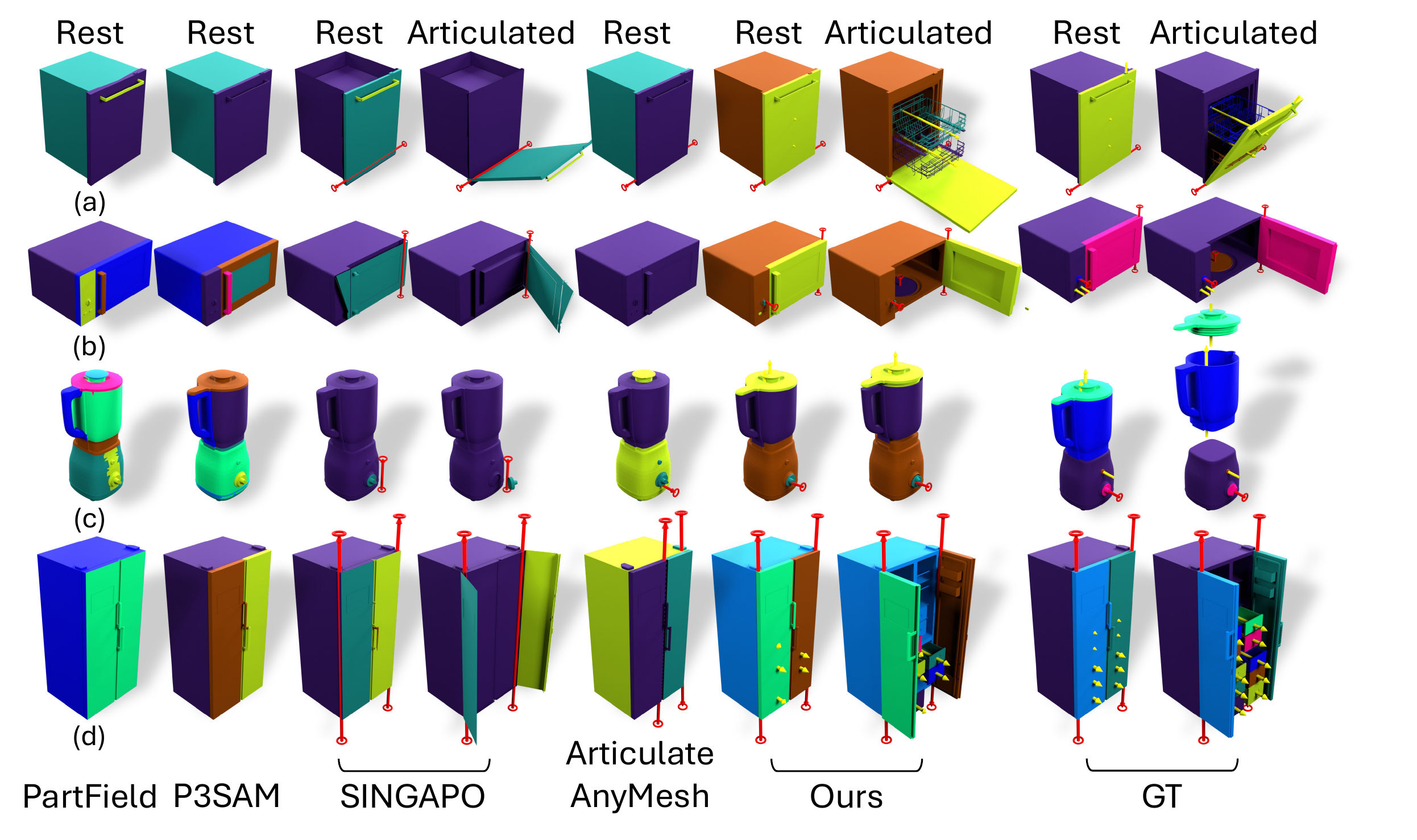}
\vspace{-2em}
\caption{
\textbf{Qualitative comparison} on the Lightwheel benchmark.
Compared with \emph{semantic} part segmentation methods (PartField~\cite{liu2025partfield} and P3SAM~\cite{ma2025p3sam}), the segments of \method better correspond to \emph{articulated} parts.
In contrast to Articulate AnyMesh~\cite{qiu2025articulate} and SINGAPO~\cite{liusingapo}, which fail to predict internal structures invisible from external views (\eg, a, b and d), \method recovers most of these parts and faithfully infers their motion constraints.
}%
\label{fig:qualitative-comp}
\vspace{-0.5em}
\end{figure}

%% file: tables/main_articulation.tex
\begin{table}[t]
\centering
\small
\setlength{\tabcolsep}{3pt}
\renewcommand{\arraystretch}{1.05}
\resizebox{\linewidth}{!}{%
\begin{tabular}{@{}lcccccc@{}}
\toprule
 & \multicolumn{3}{c}{\textbf{Lightwheel}} & \multicolumn{3}{c}{\textbf{PartNet-Mobility}} \\
\cmidrule(lr){2-4}\cmidrule(lr){5-7}
\textbf{Method} & gIoU$\uparrow$ & PC$\downarrow$ & OC$\downarrow$
& gIoU$\uparrow$ & PC$\downarrow$ & OC$\downarrow$ \\

\midrule
Naive Baseline & 0.016 & 0.293 & 0.017 & 0.296 & 0.216 & 0.027 \\
SINGAPO~\cite{liusingapo}     & -0.121 & 0.299 & 0.011 & -- & -- & -- \\
SINGAPO~\cite{liusingapo} (1@10)  & -0.100 & 0.238 & 0.012 & -- & -- & -- \\
\method (ours)          & \secondbest{0.165} & \best{0.200} & \best{0.008} & \secondbest{0.842} & \secondbest{0.024} & \secondbest{0.003}\\
\midrule
SINGAPO~\cite{liusingapo}$^\dagger$     & -0.102 & 0.329 & 0.018 &  0.255 & 0.184 & 0.046 \\
SINGAPO~\cite{liusingapo} (1@10)$^\dagger$    & -0.056 & 0.261 & 0.019 &  0.264 & 0.168 & 0.041 \\
Articulate AnyMesh~\cite{qiu2025articulate}$^\dagger$    & 0.158 & 0.237 & 0.010 & 0.378 & 0.251 & 0.022 \\
\method (ours)$^\dagger$     & \best{0.305} & \secondbest{0.208} & \secondbest{0.009} & \best{0.843} & \best{0.022} & \best{0.003} \\
\bottomrule
\end{tabular}%
}
\caption{
\textbf{Results evaluated using fully articulated geometries}, where every part is moved to its maximum extent.
We report part-wise gIoU and PC as in \cref{tab:rest-comparison}, and whole-object Chamfer distance (OC), which measures the overall bidirectional Chamfer distance between the \emph{whole} predicted geometry and its GT counterpart.
Notations $^\dagger$, --, 1@10: same as \cref{tab:rest-comparison}.
Colors: \best{best} and \secondbest{second best}.
}%
\vspace{-1em}
\label{tab:full-comparison}
\end{table}

%% file: tables/ablation_lightwheel.tex
\begin{table}[t]
\centering
\small
\setlength{\tabcolsep}{3pt}
\resizebox{\linewidth}{!}{%
\begin{tabular}{@{}lccccccccc@{}}
\toprule
 & \multirow{2}{*}{\textbf{Data}} & \multirow{2}{*}{\shortstack{\textbf{Part-} \\ \textbf{Field}}} & \multirow{2}{*}{\shortstack{\textbf{Axis} \\ \textbf{over-} \\ \textbf{param.}}} & \multicolumn{3}{c}{\textbf{Articulated state}} & \multicolumn{3}{c}{\textbf{Rest state}} \\
\cmidrule(lr){5-7}\cmidrule(lr){8-10}
 &  &  &  & gIoU$\uparrow$ & PC$\downarrow$ & OC$\downarrow$
 & gIoU$\uparrow$ & PC$\downarrow$ & mIoU $\uparrow$ \\
\midrule
$\mathbb{A}$ & PM+GRS & $\checkmark$ & $\checkmark$
& \best{0.259} &  \best{0.215} & \best{0.008} & \best{0.286} &  \secondbest{0.184} & \best{0.551} \\
$\mathbb{B}$ & PM & $\checkmark$ & $\checkmark$
& 0.231 & 0.227 & 0.013 & 0.252 & \best{0.175} & \secondbest{0.504} \\
$\mathbb{C}$ & PM+GRS & $\checkmark$ & $\times$
& \secondbest{0.236} & \secondbest{0.224} & \best{0.008} & \secondbest{0.262}  & 0.186 & 0.411 \\
$\mathbb{D}$ & PM+GRS & $\times$ & $\checkmark$
& 0.174 & 0.245 & \secondbest{0.009} & 0.195 & 0.201 & 0.491 \\
\bottomrule
\end{tabular}%
}
\caption{\textbf{Ablations} on training data (PM: PartNet-Mobility, GRS: GRSCenes), revolute axis over-parameterization and pre-trained PartField features, evaluated on (a subset of) the Lightwheel~\cite{simready2025} benchmark.
Colors: \best{best} and \secondbest{second best}.
}%
\label{tab:ablation}
\vspace{-1em}
\end{table}

%% file: sec/5_conclusion.tex
\section{Conclusion}%
\label{sec:conclusion}
We have presented \method{}, a feed-forward approach that directly predicts all attributes of the underlying articulated structure from a single static 3D mesh.
Our approach departs from prior methods by training an end-to-end network with a simple and scalable architecture on a diverse collection of articulated 3D assets.
The resulting model is significantly faster and more accurate at recovering articulated structures than prior methods.
\method also excels on AI-generated objects, enabling the creation of articulated assets compatible with physics simulators directly from images or text prompts.

\paragraph{Limitations.}

Despite the promising results, \method{} has some limitations.
First, while \method{} generalizes well to unseen instances, it fails to recover articulated objects with very different kinematic structures from those seen during training.
Fundamentally, this is because the available training data remains several orders of magnitude smaller than datasets in other domains (\eg, LAION~\cite{schuhmann2022laion} for images or Objaverse~\cite{deitke2023objaverse, objaverseXL} for static 3D meshes).
Second, although combining \method{} with an off-the-shelf 3D generator enables the creation of diverse articulated assets, these assets often exhibit inter-part penetrations, due to both artifacts in generated meshes and imperfect motion predictions.
Enhancing the physical plausibility of the generated articulated assets (\eg, via post-training~\cite{li25dso}) to enable large-scale sim-to-real training for robotics is a promising direction for future work.

\paragraph{Acknowledgments.}

Ruining Li is supported by a Toshiba Research Studentship.
Chuanxia Zheng is supported by NTU SUG-NAP and National Research Foundation, Singapore, under its NRF Fellowship Award NRF-NRFF17-2025-0009.
Christian Rupprecht is supported by an Amazon Research Award.
This work is partially supported by the UKRI AIRR programme (ID: u5ex) and ERC CoG 101001212-UNION\@.

%% file: sec/X_suppl.tex
\maketitlesupplementary

\section{Additional Experimental Results}%
\label{sec:appendix_results}

\subsection{Additional Details of Evaluation Metrics}%
\label{sec:appendix_results_eval_details}

\paragraph{Motivation for imposing penalties on unmatched parts.}
Recall that we apply Hungarian matching to obtain the correspondence between the articulated parts predicted by the model and the ground-truth parts.
Certain parts are unmatched, as the number of predicted parts is not necessarily equal to the number of GT parts.
Unlike previous work~\cite{liusingapo, liu2024cage, lei2023nap} that simply ignores the unmatched parts, we impose a penalty (corresponding to the worst possible score for each evaluation metric, \eg, $-1$ for gIoU and $0$ for mIoU) on each unmatched part.
To motivate this metric design, in \cref{tab:sup-rest} and \cref{tab:sup-articulate}, we report the evaluation results of the non-penalizing version used in prior work.
Note that the naive baseline, which treats the entire object as a single static part, outperforms \emph{all} baseline methods in non-penalizing gIoU, PC and mIoU for both the rest state and the fully articulated state, even though human judges generally find the non-naive baselines' results (\ifarxiv \cref{fig:qualitative-comp}) \else Fig. 4 \fi more reasonable.
This is because the single part predicted by the naive baseline is matched to the object's base part, which typically occupies most of the volume and is surrounded by smaller adjacent parts, resulting in favorable metric scores.
By penalizing missed small parts, our proposed metric is more consistent with human-judged quality.

\paragraph{Motivation for the whole-object Chamfer distance.}
In \ifarxiv \cref{tab:full-comparison}, \else Tab. 2, \fi
we reported the \emph{whole-object} Chamfer distance (OC), which measures the bi-directional Chamfer distance between the entire predicted and ground-truth point clouds at the fully articulated state.
In contrast to the part-averaged metrics (\ie, gIoU and PC), OC does not rely on matching predicted parts to ground-truth parts and instead provides a more \emph{global} evaluation of both part segmentation and motion parameter estimation.
In addition, many parts move farther from the base part at the fully articulated state, thus yielding a poorer (and more appropriate) OC value for the naive baseline.

\input{tables/main_lightwheel_sup}

\subsection{Per-Category Evaluation Results}%
\label{sec:appendix_results_per_category}

\cref{tab:lightwheel-category-split} provides the category distribution of our introduced Lightwheel benchmark.
\cref{tab:quant-13cats-sidecol} presents the evaluation results on all $14$ individual categories.
While \method outperforms baseline methods on almost all categories, the performance drops significantly on out-of-distribution categories such as stand mixer and stovetop, indicating our approach is still bottlenecked by the limited diversity in the training data.

\input{tables/lightwheel_category_split}

Articulate AnyMesh~\cite{qiu2025articulate} performs better specifically on the stand mixer category.
This is because stand mixer never appears in its training set and its articulation pattern is distinct from all other training objects.
With its strong VLM-based generalization, Articulate AnyMesh handles this unseen category particularly well.
For the range hood category, objects often contain a very large base component that occupies the majority of the object's volume.
SINGAPO~\cite{liusingapo} tends to predict oversized parts, especially when the input object falls outside its training distribution.
Because the ground-truth base part is already extremely large, SINGAPO's tendency to over-predict part extents accidentally aligns with the dominant fixed region, leading to inflated IoU values for this specific category.

\input{figs/fig-failure-cases}
\subsection{Failure Cases}%
\label{sec:appendix_results_failure}

We show typical failure cases of \method in~\cref{fig:failure-cases}.
Notably, while our model is trained on a large variety of articulated objects, the training dataset remains several orders of magnitude smaller than those used by open-domain 3D generators (\ie, Objaverse~\cite{deitke2023objaverse, objaverseXL}).
As a result, \method can struggle with objects whose articulation configurations deviate significantly from those represented in our training data (\textbf{left}).
Moreover, many AI-generated 3D assets lack realistic internal structures (\textbf{right}), introducing a distribution shift (since all objects in our training set contain well-defined internal parts) that can cause our model to fail.
Enhancing the model's robustness to such noisy or unclean inputs is a valuable direction for future work.

\input{tables/lightwheel_category_wise}

%% file: tables/main_lightwheel_sup.tex
\newcommand{\tabstd}[1]{\,\mbox{\tiny$\pm$\,#1}}

\begin{table}[t]
\centering
\small
\setlength{\tabcolsep}{4pt}
\renewcommand{\arraystretch}{1.05}
\resizebox{\linewidth}{!}{%
\begin{tabular}{@{}lcccccc@{}}
\toprule
 & \multicolumn{3}{c}{\textbf{Lightwheel}} & \multicolumn{3}{c}{\textbf{PartNet-Mobility}} \\
\cmidrule(lr){2-4}\cmidrule(lr){5-7}
\textbf{Method}
& gIoU $\uparrow$
& PC $\downarrow$
& mIoU $\uparrow$
& gIoU $\uparrow$
& PC $\downarrow$
& mIoU $\uparrow$ \\
\midrule
\ \underline{\emph{Without Penalty}} & & & & & & \\
Naive Baseline & \secondbest{0.687} & \secondbest{0.019} & \secondbest{0.687} & \best{0.897} & 0.003 & \best{0.897} \\
SINGAPO~\cite{liusingapo}       & 0.184 & 0.044 & 0.370  &-- & -- & -- \\
SINGAPO~\cite{liusingapo} (1@10)        & 0.250 & 0.033 & 0.404 &-- & -- & -- \\
\method (ours)           & 0.486\tabstd{0.007} & 0.028\tabstd{0.001} & 0.541\tabstd{0.005} & 0.882\tabstd{0.006} & \secondbest{0.002}\tabstd{0.000} & 0.884\tabstd{0.005}\\
\midrule
PartField$^\dagger$   & 0.092 & 0.099 & 0.268 & 0.370 & 0.040 & 0.421 \\
P3SAM$^\dagger$      & 0.421 & 0.042 & 0.523 & 0.390 & 0.023 & 0.403 \\
SINGAPO~\cite{liusingapo}$^\dagger$   & 0.278 & 0.051 & 0.394  & 0.538 & 0.008 & 0.570 \\
SINGAPO(best)~\cite{liusingapo} (1@10)$^\dagger$         & 0.407 & 0.036 & 0.476 & 0.563 & 0.006 & 0.585 \\
Articulate AnyMesh~\cite{qiu2025articulate}$^\dagger$   & 0.567 & 0.025 & 0.608 & 0.581 & 0.025 & 0.621 \\
\method (ours)$^\dagger$  & \best{0.726}\tabstd{0.052} & \best{0.010}\tabstd{0.003} & \best{0.744}\tabstd{0.047} & \secondbest{0.883}\tabstd{0.003} & \best{0.001}\tabstd{0.000} & \secondbest{0.885}\tabstd{0.003} \\
\midrule
\midrule
\ \underline{\emph{With Penalty}} & & & & & & \\
Naive Baseline & 0.018 & 0.285 & 0.413 & 0.296 & 0.210 & 0.612 \\
SINGAPO~\cite{liusingapo}        & -0.116 & 0.201 & 0.272 & -- & -- & -- \\
SINGAPO~\cite{liusingapo} (1@10)        & -0.096 & 0.190 & 0.277 & -- & -- & -- \\
\method (ours)           & \secondbest{0.183}\tabstd{0.005} & \best{0.163}\tabstd{0.001} & 0.430\tabstd{0.003} & \secondbest{0.879}\tabstd{0.006} & \secondbest{0.003}\tabstd{0.000} & \secondbest{0.883}\tabstd{0.005} \\
\midrule
PartField~\cite{liu2025partfield}$^\dagger$  & 0.079 & 0.106 & 0.264  & 0.183 & 0.123 & 0.361 \\
P3SAM~\cite{ma2025p3sam}$^\dagger$     & 0.122 & 0.177 & 0.411 & -0.116 & 0.261 & 0.267 \\
SINGAPO~\cite{liusingapo}$^\dagger$   & -0.097 & 0.234 & 0.273 & 0.262 & 0.124 & 0.468 \\
SINGAPO~\cite{liusingapo} (1@10)$^\dagger$  & -0.050 & 0.221 & 0.297 &   0.271 & 0.117 & 0.471 \\
Articulate AnyMesh~\cite{qiu2025articulate}$^\dagger$   & 0.172 & 0.190 & \secondbest{0.452} & 0.383 & 0.104 & 0.542 \\
\method (ours)$^\dagger$ & \best{0.332}\tabstd{0.034} & \secondbest{0.168}\tabstd{0.002} & \best{0.576}\tabstd{0.035} & \best{0.880}\tabstd{0.003} & \secondbest{0.003}\tabstd{0.001} & \best{0.884}\tabstd{0.003}\\
\bottomrule
\end{tabular}%
} %
\caption{\textbf{Part segmentation results at the rest state} with and without \emph{penalty} applied to unmatched parts.
The metrics without penalty do not sufficiently penalize missing small parts.
Under this protocol, the Naive Baseline outperforms \emph{all} baseline methods on \emph{all} metrics.
$^\dagger$: leveraging mesh connectivity.
1@10: reporting best out of 10 predictions.
Colors: \best{best} and \secondbest{second best}.
$\pm \sigma$: standard deviation across 10 predictions.
}%
\label{tab:sup-rest}
\end{table}

\begin{table}[t]
\centering
\small
\setlength{\tabcolsep}{4pt}
\renewcommand{\arraystretch}{1.05}
\resizebox{\linewidth}{!}{%
\begin{tabular}{@{}lcccccc@{}}
\toprule
 & \multicolumn{3}{c}{\textbf{Lightwheel}} & \multicolumn{3}{c}{\textbf{PartNet-Mobility}} \\
\cmidrule(lr){2-4}\cmidrule(lr){5-7}
\textbf{Method} & gIoU $\uparrow$ & PC $\downarrow$ & OC $\downarrow$
& gIoU $\uparrow$ & PC $\downarrow$ & OC $\downarrow$ \\
\midrule
\ \underline{\emph{Without Penalty}} & & & & & & \\
Naive Baseline & \secondbest{0.680} & \best{0.056} & 0.018 & \best{0.897} & \best{0.011} & 0.027 \\
SINGAPO~\cite{liusingapo}        & 0.178 & 0.173 & 0.011 & -- & -- & -- \\
SINGAPO~\cite{liusingapo} (1@10)      & 0.243 & 0.096 & 0.012 & -- & -- & -- \\
\method (ours)          & 0.464\tabstd{0.007} & 0.076\tabstd{0.003} & \best{0.008}\tabstd{0.000} & 0.845\tabstd{0.006} & 0.023\tabstd{0.003} & \secondbest{0.003}\tabstd{0.000}\\
\midrule
SINGAPO~\cite{liusingapo}$^\dagger$   & 0.271 & 0.179 & 0.018  & 0.523 & 0.079 & 0.046 \\
SINGAPO~\cite{liusingapo} (1@10)$^\dagger$         & 0.398 & 0.091 & 0.019 & 0.555 & 0.066 & 0.043 \\
Articulate AnyMesh~\cite{qiu2025articulate}$^\dagger$   & 0.547 & 0.084 & 0.010 & 0.577 & 0.187 & 0.022 \\
\method (ours)$^\dagger$  & \best{0.692}\tabstd{0.047} & \secondbest{0.062}\tabstd{0.004} & \secondbest{0.009}\tabstd{0.002} & \secondbest{0.846}\tabstd{0.003} & \secondbest{0.021}\tabstd{0.005} & \best{0.003}\tabstd{0.000} \\
\midrule
\midrule
\ \underline{\emph{With Penalty}} & & & & & & \\
Naive Baseline & 0.016 & 0.293 & 0.017 & 0.296 & 0.216 & 0.027 \\
SINGAPO~\cite{liusingapo}     & -0.121 & 0.299 & 0.011 & -- & -- & -- \\
SINGAPO~\cite{liusingapo} (1@10)  & -0.100 & 0.238 & 0.012 & -- & -- & -- \\
\method (ours)          & \secondbest{0.165}\tabstd{0.005} & \best{0.200}\tabstd{0.002} & \best{0.008}\tabstd{0.000} & \secondbest{0.842}\tabstd{0.005} & \secondbest{0.024}\tabstd{0.003} & \secondbest{0.003}\tabstd{0.000}\\
\midrule
SINGAPO~\cite{liusingapo}$^\dagger$     & -0.102 & 0.329 & 0.018 &  0.255 & 0.184 & 0.046 \\
SINGAPO~\cite{liusingapo} (1@10)$^\dagger$    & -0.056 & 0.261 & 0.019 &  0.264 & 0.168 & 0.041 \\
Articulate AnyMesh~\cite{qiu2025articulate}$^\dagger$    & 0.158 & 0.237 & 0.010 & 0.378 & 0.251 & 0.022 \\
\method (ours)$^\dagger$     & \best{0.305}\tabstd{0.030} & \secondbest{0.208}\tabstd{0.003} & \secondbest{0.009}\tabstd{0.002} & \best{0.843}\tabstd{0.003} & \best{0.022}\tabstd{0.005} & \best{0.003}\tabstd{0.000} \\
\bottomrule
\end{tabular}%
}
\caption{\textbf{Results evaluated using fully articulated geometries} with and without \emph{penalty} applied to unmatched parts.
The metrics without penalty do not sufficiently penalize missing small parts.
Under this protocol, the Naive Baseline outperforms \emph{all} baseline methods on \emph{all} part-wise metrics (\ie, gIoU and PC).
$^\dagger$: leveraging mesh connectivity.
1@10: reporting best out of 10 predictions.
Colors: \best{best} and \secondbest{second best}.
$\pm \sigma$: standard deviation across 10 predictions.
}%
\label{tab:sup-articulate}
\end{table}

%% file: tables/lightwheel_category_split.tex
\begin{table}[t]
    \centering
    \small
    \setlength{\tabcolsep}{6pt}
    \renewcommand{\arraystretch}{1.05}
    \begin{tabular}{@{}lc@{}}
    \toprule
    \textbf{Category} & \textbf{\# Objects} \\
    \midrule
    \cellcolor{green!15}Blender & \cellcolor{green!15}13 \\
    Coffee machine & 25 \\
    Dishwasher & 13 \\
    Electric kettle & 13 \\
    Microwave & 25 \\
    Oven & 12 \\
    \cellcolor{green!15}Range hood & \cellcolor{green!15}12 \\
    Refrigerator & 25 \\
    Sink & 25 \\
    \cellcolor{green!15}Stand mixer & \cellcolor{green!15}12 \\
    \cellcolor{green!15}Stove & \cellcolor{green!15}10 \\
    \cellcolor{green!15}Stovetop & \cellcolor{green!15}8 \\
    Toaster & 25 \\
    Toaster oven & 25 \\
    \textbf{Total} & \textbf{243} \\
    \bottomrule
    \end{tabular}
    \caption{
    \textbf{Category statistics} of our introduced Lightwheel benchmark.
    Categories highlighted in \colorbox{green!15}{green} indicate out-of-distribution categories absent from PartNet-Mobility and GRScenes.
    }
    \label{tab:lightwheel-category-split}
    \end{table}

%% file: figs/fig-failure-cases.tex
\begin{figure}[t]
\centering
\includegraphics[trim={0pt 0 50pt 0}, clip, width=\columnwidth]{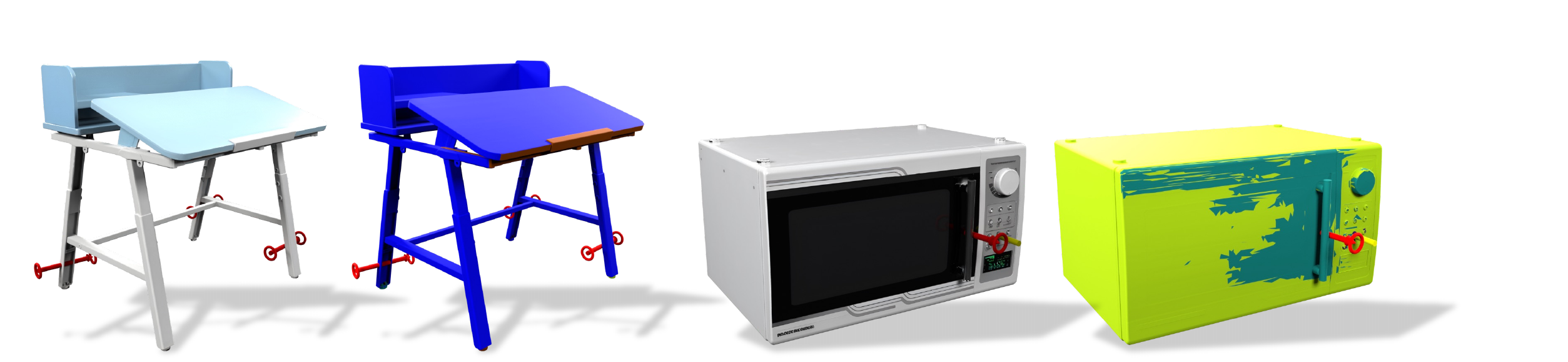}
\caption{
\textbf{Failure cases.}
Our model can struggle with objects with atypical articulation configurations (\textbf{left}) or with AI-generated shapes which lack internal structures (\textbf{right}).
}%
\label{fig:failure-cases}
\end{figure}

%% file: tables/lightwheel_category_wise.tex
\begin{table*}[ht]
\centering
\scriptsize
\setlength{\tabcolsep}{2pt}
\renewcommand{\arraystretch}{1.05}
\newcommand{\tcell}[2]{\parbox[c]{#1}{\centering #2}}

\resizebox{\linewidth}{!}{%
\begin{tabular}{@{}llccccc>{\columncolor{green!15}}c cc>{\columncolor{green!15}}c cc>{\columncolor{green!15}}c>{\columncolor{green!15}}c>{\columncolor{green!15}}c}
\toprule

\multicolumn{2}{c}{} &
\multicolumn{14}{c}{\textbf{Categories}} \\
\cmidrule(lr){3-16}
\textbf{Methods} & \textbf{Metrics}  &
    \tcell{1cm}{ Micro-\\wave } &
    \tcell{1cm}{ Dish-\\washer } &
    \tcell{1cm}{ Sink } &
    \tcell{1cm}{ Electric\\Kettle } &
    \tcell{1cm}{ Refri-\\gerator } &
    \tcell{1cm}{ Blender } &
    \tcell{1cm}{ Oven } &
    \tcell{1cm}{ Toaster } &
    \tcell{1cm}{ Stand\\Mixer } &
    \tcell{1cm}{ Toaster\\Oven } &
    \tcell{1cm}{ Coffee\\Machine } &
    \tcell{1cm}{ Stove } &
    \tcell{1cm}{ Stovetop } &
    \tcell{1cm}{ Range\\Hood } \\
    
    \midrule
    P3SAM$^\dagger$ 
& mIoU (rest) & 0.492       & 0.411       & 0.482       & \secondbest{0.637}       & \secondbest{0.498}       & \secondbest{0.514}       & 0.364       & 0.359       & \best{0.569}       & 0.395       & \secondbest{0.393}       & 0.253       & \best{0.377}       & 0.097       \\
& gIoU (rest) & 0.252       & 0.156       & 0.162       & \best{0.515}       & 0.229       & \secondbest{0.282}       & 0.092       & 0.046       & \secondbest{0.365}       & \best{0.135}       & \best{0.119}       & -0.201      & \secondbest{-0.034}       & -0.223      \\
& PC (rest)   & \best{0.103}       & 0.144       & 0.228       & \secondbest{0.110}       & 0.162       & 0.116       & \secondbest{0.170}       & \secondbest{0.187}       & \secondbest{0.146}       & \secondbest{0.152}       & \secondbest{0.156}       & 0.212       & 0.183       & \secondbest{0.233}       \\

    \midrule
    PartField$^\dagger$
    & mIoU (rest) & 0.408       & 0.273       & 0.406       & 0.320       & 0.235       & 0.330       & 0.202       & 0.192       & \secondbest{0.445}       & 0.187       & 0.205       & 0.170       & 0.222       & 0.056       \\
    & gIoU (rest) & 0.254       & 0.087       & 0.309       & 0.218       & 0.068       & 0.141       & -0.029      & -0.026      & \best{0.367}       & -0.056      & \secondbest{0.048}       & -0.166      & \secondbest{-0.023}       & -0.204      \\
    & PC (rest)   & 0.103       & \secondbest{0.104}       & 0.062       & \best{0.090}       & \best{0.070}       & \best{0.072}       & \best{0.129}       & \best{0.146}       & \best{0.063}       & \best{0.125}       & \best{0.083}       & 0.154       & \secondbest{0.134}       & \best{0.144}       \\
    
    \midrule
    SINGAPO  
& mIoU (rest)    & 0.337       & 0.346       & 0.423       & 0.304       & 0.168       & 0.050       & 0.220       & 0.248       & 0.071       & 0.193       & 0.301       & 0.225       & 0.176       & \secondbest{0.443}       \\
& gIoU (rest)    & 0.003       & -0.080      & 0.135       & -0.286      & -0.126      & -0.328      & -0.162      & -0.172      & -0.335      & -0.180      & -0.144      & -0.190      & -0.234      & \secondbest{0.000}       \\
& gIoU (high)    & 0.001       & -0.084      & 0.135       & -0.328      & -0.129      & -0.331      & -0.166      & -0.176      & -0.337      & -0.182      & -0.146      & -0.196      & -0.241      & \secondbest{-0.001}       \\
& PC (rest)      & 0.138       & 0.196       & 0.197       & 0.313       & 0.126       & 0.236       & 0.187       & 0.208       & 0.326       & 0.192       & 0.273       & 0.163       & \best{0.109}       & 0.259       \\
& PC (high)      & 0.362       & 0.401       & 0.220       & 0.593       & 0.176       & 0.318       & 0.299       & 0.281       & 0.573       & 0.293       & 0.342       & 0.205       & \best{0.134}       & 0.284       \\
& OC (high)      & 0.016       & \secondbest{0.013}       & 0.002       & 0.059       & 0.007       & 0.026       & 0.008       & 0.007       & 0.040       & 0.009       & 0.013       & 0.004       & 0.001       & 0.003       \\

    \midrule
    SINGAPO$^\dagger$ 
& mIoU (rest)    & 0.392       & 0.331       & 0.056       & 0.490       & 0.209       & 0.102       & 0.251       & 0.283       & 0.067       & 0.229       & 0.376       & 0.279       & 0.254       & \best{0.515}       \\
& gIoU (rest)    & 0.080       & -0.075      & -0.262      & 0.072       & -0.040      & -0.271      & -0.139      & -0.114      & -0.349      & -0.160      & -0.057      & -0.132      & -0.114      & \best{0.089}       \\
& gIoU (high)    & 0.079       & -0.077      & -0.262      & 0.008       & -0.047      & -0.271      & -0.145      & -0.117      & -0.349      & -0.164      & -0.057      & -0.136      & -0.117      & \best{0.089}       \\
& PC (rest)      & 0.171       & 0.225       & 0.278       & 0.288       & 0.124       & 0.261       & 0.240       & 0.241       & 0.355       & 0.247       & 0.289       & 0.225       & 0.185       & 0.265       \\
& PC (high)      & 0.435       & 0.425       & 0.282       & 0.338       & 0.182       & 0.341       & 0.366       & 0.326       & 0.625       & 0.364       & 0.329       & 0.279       & 0.202       & 0.283       \\
& OC (high)      & 0.019       & 0.013       & 0.059       & 0.054       & 0.007       & 0.025       & 0.009       & 0.007       & \secondbest{0.040}       & 0.011       & 0.013       & 0.005       & 0.001       & 0.002       \\
    
    \midrule
    \multirow{2}{*}{\parbox[t]{1.5cm}{Articulate \\ AnyMesh~\cite{qiu2025articulate}$^\dagger$}} 
& mIoU (rest)    & 0.671       & 0.583       & \secondbest{0.493}       & 0.408       & 0.362       & 0.503       & 0.432       & \secondbest{0.441}       & 0.425       & 0.387       & 0.340       & 0.369       & 0.289       & 0.236       \\
& gIoU (rest)    & 0.434       & 0.283       & \secondbest{0.402}       & 0.092       & 0.081       & 0.275       & 0.122       & \secondbest{0.130}       & 0.210       & 0.063       & 0.000       & -0.033      & -0.129      & -0.104      \\
& gIoU (high)    & 0.405       & 0.265       & \secondbest{0.394}       & 0.053       & 0.077       & \secondbest{0.247}       & 0.107       & \secondbest{0.120}       & \best{0.178}       & 0.057       & \secondbest{-0.005}       & -0.040      & -0.129      & -0.104      \\
& PC (rest)      & 0.147       & 0.218       & \secondbest{0.057}       & 0.218       & 0.157       & 0.133       & 0.235       & 0.222       & 0.155       & 0.236       & 0.210       & 0.281       & 0.273       & 0.280       \\
& PC (high)      & 0.216       & 0.235       & \best{0.068}       & 0.472       & 0.215       & 0.182       & 0.279       & 0.258       & \best{0.451}       & 0.297       & \best{0.227}       & 0.292       & 0.289       & \secondbest{0.279}       \\
& OC (high)      & 0.011       & \best{0.003}       & \best{0.001}       & 0.064       & 0.011       & \secondbest{0.021}       & 0.011       & 0.009       & 0.040       & 0.016       & 0.014       & \secondbest{0.002}       & \best{0.000}       & \best{0.000}       \\
    
    \midrule
    \multirow{2}{*}{\parbox[t]{1.5cm}{\method \\ (ours)}} 
& mIoU (rest)    & \secondbest{0.678}       & \secondbest{0.715}       & 0.369       & 0.503       & 0.488       & 0.404       & \secondbest{0.491}       & 0.393       & 0.267       & \secondbest{0.444}       & 0.256       & \secondbest{0.392}       & 0.222       & 0.208       \\
& gIoU (rest)    & \secondbest{0.472}       & \secondbest{0.593}       & 0.259       & 0.218       & \secondbest{0.287}       & 0.196       & \secondbest{0.199}       & 0.109       & -0.068      & 0.102       & -0.144      & \secondbest{0.196}       & -0.041      & -0.189      \\
& gIoU (high)    & \secondbest{0.439}       & \secondbest{0.554}       & 0.257       & \secondbest{0.167}       & \secondbest{0.265}       & 0.173       & \best{0.180}       & 0.098       & \secondbest{-0.083}       & \secondbest{0.086}       & -0.147      & \secondbest{0.186}       & \secondbest{-0.042}       & -0.191      \\
& PC (rest)      & 0.130       & \best{0.088}       & 0.092       & 0.208       & 0.120       & \secondbest{0.111}       & 0.212       & 0.188       & 0.251       & 0.242       & 0.224       & \best{0.106}       & 0.166       & 0.328       \\
& PC (high)      & \secondbest{0.123}       & \best{0.172}       & \secondbest{0.140}       & \best{0.274}       & \secondbest{0.123}       & \best{0.165}       & \best{0.219}       & \best{0.197}       & 0.509       & \best{0.244}       & \secondbest{0.267}       & \best{0.110}       & \secondbest{0.179}       & 0.338       \\
& OC (high)      & \secondbest{0.001}       & 0.024       & \secondbest{0.002}       & \best{0.018}       & \best{0.001}       & \best{0.019}       & \best{0.005}       & \best{0.001}       & 0.043       & \best{0.002}       & \best{0.012}       & 0.002       & 0.003       & \secondbest{0.002}       \\
    
    \midrule
    
    \multirow{2}{*}{\parbox[t]{1.5cm}{\method \\ (ours)$^\dagger$}} 
& mIoU (rest)    & \best{0.802}       & \best{0.811}       & \best{0.508}       & \best{0.674}       & \best{0.633}       & \best{0.708}       & \best{0.521}       & \best{0.509}       & 0.284       & \best{0.490}       & \best{0.435}       & \best{0.543}       & \secondbest{0.373}       & 0.303       \\
& gIoU (rest)    & \best{0.632}       & \best{0.689}       & \best{0.425}       & \secondbest{0.390}       & \best{0.498}       & \best{0.501}       & \best{0.203}       & \best{0.199}       & -0.071      & \secondbest{0.134}       & 0.038       & \best{0.292}       & \best{-0.007}       & -0.088      \\
& gIoU (high)    & \best{0.580}       & \best{0.640}       & \best{0.423}       & \best{0.330}       & \best{0.452}       & \best{0.456}       & \secondbest{0.179}       & \best{0.181}       & -0.083      & \best{0.110}       & \best{0.033}       & \best{0.274}       & \best{-0.012}       & -0.088      \\
& PC (rest)      & \secondbest{0.115}       & \secondbest{0.088}       & \best{0.047}       & 0.207       & \secondbest{0.072}       & 0.121       & 0.234       & 0.214       & 0.281       & 0.252       & 0.278       & \secondbest{0.141}       & 0.195       & 0.273       \\
& PC (high)      & \best{0.116}       & \secondbest{0.177}       & 0.158       & \secondbest{0.276}       & \best{0.075}       & \secondbest{0.171}       & \secondbest{0.242}       & \secondbest{0.220}       & \secondbest{0.482}       & \secondbest{0.254}       & 0.321       & \secondbest{0.147}       & 0.203       & \best{0.276}       \\
& OC (high)      & \best{0.000}       & 0.027       & 0.003       & \secondbest{0.018}       & \secondbest{0.001}       & \secondbest{0.019}       & \secondbest{0.005}       & \secondbest{0.001}       & \best{0.038}       & \secondbest{0.002}       & \secondbest{0.012}       & \best{0.001}       & \secondbest{0.001}       & 0.002       \\

\bottomrule
\end{tabular}
}
\caption{
\textbf{Per-category evaluation results} against baselines on the Lightwheel dataset.
We report the results both at the rest state (\textbf{rest}) and at the fully articulated state (\textbf{high}).
The categories are ranked based on gIoU (rest) averaged over all methods (left is higher).
$^\dagger$: leveraging mesh connectivity.
Categories highlighted in \colorbox{green!15}{green} indicate out-of-distribution categories absent from the training datasets PartNet-Mobility and GRScenes.
Colors: \best{best} and \secondbest{second best}.
}%

\label{tab:quant-13cats-sidecol}
\end{table*}

%% file: main.bbl
\begin{thebibliography}{69}
\providecommand{\natexlab}[1]{#1}
\providecommand{\url}[1]{\texttt{#1}}
\expandafter\ifx\csname urlstyle\endcsname\relax
  \providecommand{\doi}[1]{doi: #1}\else
  \providecommand{\doi}{doi: \begingroup \urlstyle{rm}\Url}\fi

\bibitem[Abdelreheem et~al.(2023)Abdelreheem, Skorokhodov, Ovsjanikov, and
  Wonka]{abdelreheem2023satr}
Ahmed Abdelreheem, Ivan Skorokhodov, Maks Ovsjanikov, and Peter Wonka.
\newblock Satr: Zero-shot semantic segmentation of 3d shapes.
\newblock In \emph{ICCV}, 2023.

\bibitem[Arnaud et~al.(2025)Arnaud, McVay, Martin, Majumdar, Jatavallabhula,
  Thomas, Partsey, Dugas, Gejji, Sax, et~al.]{arnaud2025locate}
Sergio Arnaud, Paul McVay, Ada Martin, Arjun Majumdar, Krishna~Murthy
  Jatavallabhula, Phillip Thomas, Ruslan Partsey, Daniel Dugas, Abha Gejji,
  Alexander Sax, et~al.
\newblock {Locate 3D}: Real-world object localization via self-supervised
  learning in 3d.
\newblock In \emph{ICML}, 2025.

\bibitem[Carion et~al.(2020)Carion, Massa, Synnaeve, Usunier, Kirillov, and
  Zagoruyko]{carion20end-to-end}
Nicolas Carion, Francisco Massa, Gabriel Synnaeve, Nicolas Usunier, Alexander
  Kirillov, and Sergey Zagoruyko.
\newblock End-to-end object detection with transformers.
\newblock In \emph{{ECCV}}, 2020.

\bibitem[Chen et~al.(2025{\natexlab{a}})Chen, Liu, Wei, Su, and
  Liu]{chen2025freeart3d}
Chuhao Chen, Isabella Liu, Xinyue Wei, Hao Su, and Minghua Liu.
\newblock {FreeArt3D}: Training-free articulated object generation using 3d
  diffusion.
\newblock In \emph{SIGGRAPH Asia}, 2025{\natexlab{a}}.

\bibitem[Chen et~al.(2025{\natexlab{b}})Chen, Lan, Chen, and
  Pan]{chen2025ArtiLatent}
Honghua Chen, Yushi Lan, Yongwei Chen, and Xingang Pan.
\newblock {ArtiLatent}: Realistic articulated 3d object generation via
  structured latents.
\newblock In \emph{SIGGRAPH Asia}, 2025{\natexlab{b}}.

\bibitem[Chen et~al.(2025{\natexlab{c}})Chen, Zhang, Liang, Luo, Li, Liu, Li,
  Long, Feng, and Tan]{chen2025dora}
Rui Chen, Jianfeng Zhang, Yixun Liang, Guan Luo, Weiyu Li, Jiarui Liu, Xiu Li,
  Xiaoxiao Long, Jiashi Feng, and Ping Tan.
\newblock Dora: Sampling and benchmarking for 3d shape variational
  auto-encoders.
\newblock In \emph{CVPR}, 2025{\natexlab{c}}.

\bibitem[Deemos(2024)]{deemos24rodin}
Deemos.
\newblock Rodin text-to-{3D} gen-1 (0525) v0.5, 2024.

\bibitem[Deitke et~al.(2023{\natexlab{a}})Deitke, Liu, Wallingford, Ngo,
  Michel, Kusupati, Fan, Laforte, Voleti, Gadre, VanderBilt, Kembhavi,
  Vondrick, Gkioxari, Ehsani, Schmidt, and Farhadi]{objaverseXL}
Matt Deitke, Ruoshi Liu, Matthew Wallingford, Huong Ngo, Oscar Michel, Aditya
  Kusupati, Alan Fan, Christian Laforte, Vikram Voleti, Samir~Yitzhak Gadre,
  Eli VanderBilt, Aniruddha Kembhavi, Carl Vondrick, Georgia Gkioxari, Kiana
  Ehsani, Ludwig Schmidt, and Ali Farhadi.
\newblock {Objaverse-XL}: A universe of 10m+ 3d objects.
\newblock In \emph{NeurIPS}, 2023{\natexlab{a}}.

\bibitem[Deitke et~al.(2023{\natexlab{b}})Deitke, Schwenk, Salvador, Weihs,
  Michel, VanderBilt, Schmidt, Ehsani, Kembhavi, and
  Farhadi]{deitke2023objaverse}
Matt Deitke, Dustin Schwenk, Jordi Salvador, Luca Weihs, Oscar Michel, Eli
  VanderBilt, Ludwig Schmidt, Kiana Ehsani, Aniruddha Kembhavi, and Ali
  Farhadi.
\newblock {Objaverse}: A universe of annotated 3d objects.
\newblock In \emph{CVPR}, 2023{\natexlab{b}}.

\bibitem[Deng et~al.(2025)Deng, Zhang, Geng, Wu, and Wu]{deng2025anymate}
Yufan Deng, Yuhao Zhang, Chen Geng, Shangzhe Wu, and Jiajun Wu.
\newblock Anymate: A dataset and baselines for learning 3d object rigging.
\newblock In \emph{SIGGRAPH}, 2025.

\bibitem[Gao et~al.(2025)Gao, Siddiqui, Li, and Dai]{gao2025meshart}
Daoyi Gao, Yawar Siddiqui, Lei Li, and Angela Dai.
\newblock {MeshArt}: Generating articulated meshes with structure-guided
  transformers.
\newblock In \emph{CVPR}, 2025.

\bibitem[Geng et~al.(2023)Geng, Xu, Zhao, Xu, Yi, Huang, and
  Wang]{Geng_2023_CVPR}
Haoran Geng, Helin Xu, Chengyang Zhao, Chao Xu, Li Yi, Siyuan Huang, and He
  Wang.
\newblock {GAPartNet}: Cross-category domain-generalizable object perception
  and manipulation via generalizable and actionable parts.
\newblock In \emph{CVPR}, 2023.

\bibitem[Goyal et~al.(2025)Goyal, Petrov, Andrews, Ben-Shabat, Liu, and
  Kalogerakis]{goyal2025geopard}
Pradyumn Goyal, Dmitry Petrov, Sheldon Andrews, Yizhak Ben-Shabat,
  Hsueh-Ti~Derek Liu, and Evangelos Kalogerakis.
\newblock Geopard: Geometric pretraining for articulation prediction in 3d
  shapes.
\newblock In \emph{ICCV}, 2025.

\bibitem[Ho et~al.(2020)Ho, Jain, and Abbeel]{ho20denoising}
Jonathan Ho, Ajay Jain, and Pieter Abbeel.
\newblock Denoising diffusion probabilistic models.
\newblock In \emph{{NeurIPS}}, 2020.

\bibitem[Hunyuan3D et~al.(2025)Hunyuan3D, Yang, Yang, Feng, Huang, Zhang, He,
  Luo, Liu, Zhao, Lin, Lai, Yang, Shi, Zhao, Zhang, Yan, Wang, Liu, Zhang,
  Chen, Dong, Jia, Cai, Yu, Tang, Guo, Yu, Zhang, Ye, He, Wu, Wei, Zhang, Tan,
  Sun, Niu, Huang, Zheng, Liu, Chen, Yuan, Yang, Liu, Zhu, Chen, Liu, Wang,
  Liu, Linus, Jiang, Huang, and Guo]{hunyuan3d25hunyuan3d}
Team Hunyuan3D, Shuhui Yang, Mingxin Yang, Yifei Feng, Xin Huang, Sheng Zhang,
  Zebin He, Di Luo, Haolin Liu, Yunfei Zhao, Qingxiang Lin, Zeqiang Lai,
  Xianghui Yang, Huiwen Shi, Zibo Zhao, Bowen Zhang, Hongyu Yan, Lifu Wang,
  Sicong Liu, Jihong Zhang, Meng Chen, Liang Dong, Yiwen Jia, Yulin Cai, Jiaao
  Yu, Yixuan Tang, Dongyuan Guo, Junlin Yu, Hao Zhang, Zheng Ye, Peng He,
  Runzhou Wu, Shida Wei, Chao Zhang, Yonghao Tan, Yifu Sun, Lin Niu, Shirui
  Huang, Bojian Zheng, Shu Liu, Shilin Chen, Xiang Yuan, Xiaofeng Yang, Kai
  Liu, Jianchen Zhu, Peng Chen, Tian Liu, Di Wang, Yuhong Liu, Linus, Jie
  Jiang, Jingwei Huang, and Chunchao Guo.
\newblock {Hunyuan3D} 2.1: From images to high-fidelity {3D} assets with
  production-ready {PBR} material.
\newblock \emph{arXiv}, 2506.15442, 2025.

\bibitem[Jakab et~al.(2024)Jakab, Li, Wu, Rupprecht, and
  Vedaldi]{jakab24farm3d}
Tomas Jakab, Ruining Li, Shangzhe Wu, Christian Rupprecht, and Andrea Vedaldi.
\newblock {Farm3D}: Learning articulated {3D} animals by distilling {2D}
  diffusion.
\newblock In \emph{{3DV}}, 2024.

\bibitem[Jiang et~al.(2022)Jiang, Hsu, and Zhu]{jiang2022ditto}
Zhenyu Jiang, Cheng-Chun Hsu, and Yuke Zhu.
\newblock Ditto: Building digital twins of articulated objects from
  interaction.
\newblock In \emph{CVPR}, 2022.

\bibitem[Kerbl et~al.(2023)Kerbl, Kopanas, Leimk{\"u}hler, and
  Drettakis]{kerbl233d-gaussian}
Bernhard Kerbl, Georgios Kopanas, Thomas Leimk{\"u}hler, and George Drettakis.
\newblock {3D} {Gaussian Splatting} for real-time radiance field rendering.
\newblock \emph{{SIGGRAPH}}, 42\penalty0 (4), 2023.

\bibitem[Kirillov et~al.(2023)Kirillov, Mintun, Ravi, Mao, Rolland, Gustafson,
  Xiao, Whitehead, Berg, Lo, et~al.]{kirillov2023segment}
Alexander Kirillov, Eric Mintun, Nikhila Ravi, Hanzi Mao, Chloe Rolland, Laura
  Gustafson, Tete Xiao, Spencer Whitehead, Alexander~C Berg, Wan-Yen Lo, et~al.
\newblock Segment anything.
\newblock In \emph{ICCV}, 2023.

\bibitem[Le et~al.(2025)Le, Xie, Liang, Wang, Yang, Ma, Vedder, Krishna,
  Jayaraman, and Eaton]{le2024articulate}
Long Le, Jason Xie, William Liang, Hung-Ju Wang, Yue Yang, Yecheng~Jason Ma,
  Kyle Vedder, Arjun Krishna, Dinesh Jayaraman, and Eric Eaton.
\newblock {Articulate-Anything}: Automatic modeling of articulated objects via
  a vision-language foundation model.
\newblock In \emph{ICLR}, 2025.

\bibitem[Lei et~al.(2023)Lei, Deng, Shen, Guibas, and Daniilidis]{lei2023nap}
Jiahui Lei, Congyue Deng, William~B Shen, Leonidas~J Guibas, and Kostas
  Daniilidis.
\newblock Nap: Neural 3d articulation prior.
\newblock In \emph{NeurIPS}, 2023.

\bibitem[Li et~al.(2023)Li, Zhang, Wong, Gokmen, Srivastava,
  Mart{\'\i}n-Mart{\'\i}n, Wang, Levine, Lingelbach, Sun,
  et~al.]{li2023behavior}
Chengshu Li, Ruohan Zhang, Josiah Wong, Cem Gokmen, Sanjana Srivastava, Roberto
  Mart{\'\i}n-Mart{\'\i}n, Chen Wang, Gabrael Levine, Michael Lingelbach,
  Jiankai Sun, et~al.
\newblock {BEHAVIOR-1K}: A human-centered, embodied ai benchmark with 1,000
  everyday activities and realistic simulation.
\newblock In \emph{CoRL}, 2023.

\bibitem[Li et~al.(2022)Li, Zhang, Zhang, Yang, Li, Zhong, Wang, Yuan, Zhang,
  Hwang, Chang, and Gao]{li2022groundedlanguageimagepretraining}
Liunian~Harold Li, Pengchuan Zhang, Haotian Zhang, Jianwei Yang, Chunyuan Li,
  Yiwu Zhong, Lijuan Wang, Lu Yuan, Lei Zhang, Jenq-Neng Hwang, Kai-Wei Chang,
  and Jianfeng Gao.
\newblock Grounded language-image pre-training.
\newblock In \emph{CVPR}, 2022.

\bibitem[Li et~al.(2024{\natexlab{a}})Li, Zheng, Rupprecht, and
  Vedaldi]{li24dragapart}
Ruining Li, Chuanxia Zheng, Christian Rupprecht, and Andrea Vedaldi.
\newblock {DragAPart}: Learning a part-level motion prior for articulated
  objects.
\newblock In \emph{ECCV}, 2024{\natexlab{a}}.

\bibitem[Li et~al.(2025{\natexlab{a}})Li, Zheng, Rupprecht, and
  Vedaldi]{li25dso}
Ruining Li, Chuanxia Zheng, Christian Rupprecht, and Andrea Vedaldi.
\newblock {DSO}: Aligning {3D} generators with simulation feedback for physical
  soundness.
\newblock In \emph{ICCV}, 2025{\natexlab{a}}.

\bibitem[Li et~al.(2025{\natexlab{b}})Li, Zheng, Rupprecht, and
  Vedaldi]{li25puppet-master}
Ruining Li, Chuanxia Zheng, Christian Rupprecht, and Andrea Vedaldi.
\newblock Puppet-master: Scaling interactive video generation as a motion prior
  for part-level dynamics.
\newblock In \emph{ICCV}, 2025{\natexlab{b}}.

\bibitem[Li et~al.(2024{\natexlab{b}})Li, Litvak, Li, Zhang, Jakab, Rupprecht,
  Wu, Vedaldi, and Wu]{li24learning}
Zizhang Li, Dor Litvak, Ruining Li, Yunzhi Zhang, Tomas Jakab, Christian
  Rupprecht, Shangzhe Wu, Andrea Vedaldi, and Jiajun Wu.
\newblock Learning the {3D} fauna of the {Web}.
\newblock In \emph{CVPR}, 2024{\natexlab{b}}.

\bibitem[Lian et~al.(2025)Lian, Yu, Liang, Wang, Luo, Chen, Zhou, Tang, Xu,
  Lyu, et~al.]{lian2025infinite}
Xinyu Lian, Zichao Yu, Ruiming Liang, Yitong Wang, Li~Ray Luo, Kaixu Chen,
  Yuanzhen Zhou, Qihong Tang, Xudong Xu, Zhaoyang Lyu, et~al.
\newblock Infinite mobility: Scalable high-fidelity synthesis of articulated
  objects via procedural generation.
\newblock \emph{arXiv preprint arXiv:2503.13424}, 2025.

\bibitem[{Lightwheel}(2025)]{simready2025}
{Lightwheel}.
\newblock Simready: Simulation-ready 3d assets.
\newblock \url{https://simready.com/}, 2025.
\newblock Accessed: 2025.

\bibitem[Liu et~al.(2025{\natexlab{a}})Liu, Xu, Wang, Tan, Xu, Wang, Su, and
  Shi]{liu2025riganything}
Isabella Liu, Zhan Xu, Yifan Wang, Hao Tan, Zexiang Xu, Xiaolong Wang, Hao Su,
  and Zifan Shi.
\newblock {RigAnything}: Template-free autoregressive rigging for diverse 3d
  assets.
\newblock \emph{ACM TOG}, 44\penalty0 (4), 2025{\natexlab{a}}.

\bibitem[Liu et~al.(2023{\natexlab{a}})Liu, Mahdavi-Amiri, and
  Savva]{liu2023paris}
Jiayi Liu, Ali Mahdavi-Amiri, and Manolis Savva.
\newblock {PARIS}: Part-level reconstruction and motion analysis for
  articulated objects.
\newblock In \emph{ICCV}, 2023{\natexlab{a}}.

\bibitem[Liu et~al.(2024)Liu, Tam, Mahdavi-Amiri, and Savva]{liu2024cage}
Jiayi Liu, Hou In~Ivan Tam, Ali Mahdavi-Amiri, and Manolis Savva.
\newblock {CAGE}: Controllable articulation generation.
\newblock In \emph{CVPR}, 2024.

\bibitem[Liu et~al.(2025{\natexlab{b}})Liu, Iliash, Chang, Savva, and
  Amiri]{liusingapo}
Jiayi Liu, Denys Iliash, Angel~X Chang, Manolis Savva, and Ali~Mahdavi Amiri.
\newblock {SINGAPO}: Single image controlled generation of articulated parts in
  objects.
\newblock In \emph{ICLR}, 2025{\natexlab{b}}.

\bibitem[Liu et~al.(2023{\natexlab{b}})Liu, Zhu, Cai, Han, Ling, Porikli, and
  Su]{liu2023partslip}
Minghua Liu, Yinhao Zhu, Hong Cai, Shizhong Han, Zhan Ling, Fatih Porikli, and
  Hao Su.
\newblock {PartSLIP}: Low-shot part segmentation for 3d point clouds via
  pretrained image-language models.
\newblock In \emph{CVPR}, 2023{\natexlab{b}}.

\bibitem[Liu et~al.(2025{\natexlab{c}})Liu, Uy, Xiang, Su, Fidler, Sharp, and
  Gao]{liu2025partfield}
Minghua Liu, Mikaela~Angelina Uy, Donglai Xiang, Hao Su, Sanja Fidler, Nicholas
  Sharp, and Jun Gao.
\newblock {PartField}: Learning 3d feature fields for part segmentation and
  beyond.
\newblock In \emph{ICCV}, 2025{\natexlab{c}}.

\bibitem[Liu et~al.(2019)Liu, Fan, Xiang, and Pan]{liu2019rscnn}
Yongcheng Liu, Bin Fan, Shiming Xiang, and Chunhong Pan.
\newblock Relation-shape convolutional neural network for point cloud analysis.
\newblock In \emph{CVPR}, 2019.

\bibitem[Liu et~al.(2025{\natexlab{d}})Liu, Jia, Lu, Ni, Zhu, and
  Huang]{liu2025building}
Yu Liu, Baoxiong Jia, Ruijie Lu, Junfeng Ni, Song-Chun Zhu, and Siyuan Huang.
\newblock Building interactable replicas of complex articulated objects via
  gaussian splatting.
\newblock In \emph{ICLR}, 2025{\natexlab{d}}.

\bibitem[Lu et~al.(2025)Lu, Liu, Tang, Ni, Wang, Wan, Zeng, Chen, and
  Huang]{lu2025dreamart}
Ruijie Lu, Yu Liu, Jiaxiang Tang, Junfeng Ni, Yuxiang Wang, Diwen Wan, Gang
  Zeng, Yixin Chen, and Siyuan Huang.
\newblock {DreamArt}: Generating interactable articulated objects from a single
  image.
\newblock \emph{arXiv preprint arXiv:2507.05763}, 2025.

\bibitem[Ma et~al.(2025{\natexlab{a}})Ma, Li, Yan, Xu, Yang, Wang, Zhao, Guo,
  Chen, and Guo]{ma2025p3sam}
Changfeng Ma, Yang Li, Xinhao Yan, Jiachen Xu, Yunhan Yang, Chunshi Wang, Zibo
  Zhao, Yanwen Guo, Zhuo Chen, and Chunchao Guo.
\newblock {P3-SAM}: Native 3d part segmentation.
\newblock \emph{arXiv preprint arXiv:2509.06784}, 2025{\natexlab{a}}.

\bibitem[Ma et~al.(2025{\natexlab{b}})Ma, Yue, and Gkioxari]{ma20253d}
Ziqi Ma, Yisong Yue, and Georgia Gkioxari.
\newblock Find any part in 3d.
\newblock In \emph{ICCV}, 2025{\natexlab{b}}.

\bibitem[Mildenhall et~al.(2020)Mildenhall, Srinivasan, Tancik, Barron,
  Ramamoorthi, and Ng]{mildenhall20nerf:}
Ben Mildenhall, Pratul~P. Srinivasan, Matthew Tancik, Jonathan~T. Barron, Ravi
  Ramamoorthi, and Ren Ng.
\newblock {NeRF}: Representing scenes as neural radiance fields for view
  synthesis.
\newblock In \emph{{ECCV}}, 2020.

\bibitem[Mu et~al.(2021)Mu, Qiu, Kortylewski, Yuille, Vasconcelos, and
  Wang]{mu2021sdf}
Jiteng Mu, Weichao Qiu, Adam Kortylewski, Alan Yuille, Nuno Vasconcelos, and
  Xiaolong Wang.
\newblock A-sdf: Learning disentangled signed distance functions for
  articulated shape representation.
\newblock In \emph{ICCV}, 2021.

\bibitem[Qiu et~al.(2025)Qiu, Yang, Wang, Chen, Wang, Wang, Xian, and
  Gan]{qiu2025articulate}
Xiaowen Qiu, Jincheng Yang, Yian Wang, Zhehuan Chen, Yufei Wang, Tsun-Hsuan
  Wang, Zhou Xian, and Chuang Gan.
\newblock {Articulate AnyMesh}: Open-vocabulary 3d articulated objects
  modeling.
\newblock In \emph{CoRL}, 2025.

\bibitem[Raistrick et~al.(2024)Raistrick, Mei, Kayan, Yan, Zuo, Han, Wen,
  Parakh, Alexandropoulos, Lipson, et~al.]{raistrick2024infinigen}
Alexander Raistrick, Lingjie Mei, Karhan Kayan, David Yan, Yiming Zuo, Beining
  Han, Hongyu Wen, Meenal Parakh, Stamatis Alexandropoulos, Lahav Lipson,
  et~al.
\newblock {Infinigen Indoors}: Photorealistic indoor scenes using procedural
  generation.
\newblock In \emph{CVPR}, 2024.

\bibitem[Rezatofighi et~al.(2019)Rezatofighi, Tsoi, Gwak, Sadeghian, Reid, and
  Savarese]{rezatofighi2019generalized}
Hamid Rezatofighi, Nathan Tsoi, JunYoung Gwak, Amir Sadeghian, Ian Reid, and
  Silvio Savarese.
\newblock Generalized intersection over union: A metric and a loss for bounding
  box regression.
\newblock In \emph{CVPR}, 2019.

\bibitem[Schuhmann et~al.(2022)Schuhmann, Beaumont, Vencu, Gordon, Wightman,
  Cherti, Coombes, Katta, Mullis, Wortsman, et~al.]{schuhmann2022laion}
Christoph Schuhmann, Romain Beaumont, Richard Vencu, Cade Gordon, Ross
  Wightman, Mehdi Cherti, Theo Coombes, Aarush Katta, Clayton Mullis, Mitchell
  Wortsman, et~al.
\newblock Laion-5b: An open large-scale dataset for training next generation
  image-text models.
\newblock \emph{NeurIPS}, 2022.

\bibitem[Shen et~al.(2021)Shen, Xia, Li, Mart{\'\i}n-Mart{\'\i}n, Fan, Wang,
  P{\'e}rez-D’Arpino, Buch, Srivastava, Tchapmi, et~al.]{shen2021igibson}
Bokui Shen, Fei Xia, Chengshu Li, Roberto Mart{\'\i}n-Mart{\'\i}n, Linxi Fan,
  Guanzhi Wang, Claudia P{\'e}rez-D’Arpino, Shyamal Buch, Sanjana Srivastava,
  Lyne Tchapmi, et~al.
\newblock {iGibson 1.0}: a simulation environment for interactive tasks in
  large realistic scenes.
\newblock In \emph{IROS}, 2021.

\bibitem[Song et~al.(2024)Song, Wei, Foo, Lin, and Liu]{song2024reacto}
Chaoyue Song, Jiacheng Wei, Chuan~Sheng Foo, Guosheng Lin, and Fayao Liu.
\newblock Reacto: Reconstructing articulated objects from a single video.
\newblock In \emph{CVPR}, 2024.

\bibitem[Song et~al.(2025)Song, Zhang, Li, Yang, Chen, Xu, Liew, Guo, Liu,
  Feng, and Lin]{song2025magicarticulate}
Chaoyue Song, Jianfeng Zhang, Xiu Li, Fan Yang, Yiwen Chen, Zhongcong Xu,
  Jun~Hao Liew, Xiaoyang Guo, Fayao Liu, Jiashi Feng, and Guosheng Lin.
\newblock {MagicArticulate}: Make your 3d models articulation-ready.
\newblock In \emph{CVPR}, 2025.

\bibitem[Su et~al.(2025)Su, Feng, Li, Song, He, Ren, and Xu]{Su_2025_CVPR}
Jiayi Su, Youhe Feng, Zheng Li, Jinhua Song, Yangfan He, Botao Ren, and Botian
  Xu.
\newblock {ArtFormer}: Controllable generation of diverse 3d articulated
  objects.
\newblock In \emph{CVPR}, 2025.

\bibitem[Tang et~al.(2024)Tang, Zhao, Ford, Benhaim, and
  Zhang]{tang2024segment}
George Tang, William Zhao, Logan Ford, David Benhaim, and Paul Zhang.
\newblock Segment any mesh: Zero-shot mesh part segmentation via lifting
  segment anything 2 to 3d.
\newblock \emph{arXiv e-prints}, pages arXiv--2408, 2024.

\bibitem[Team(2024)]{yang2024hunyuan3d}
Tencent~Hunyuan3D Team.
\newblock {Hunyuan3D 1.0}: A unified framework for text-to-3d and image-to-3d
  generation.
\newblock \emph{arXiv preprint arXiv:2411.02293}, 2024.

\bibitem[Team(2025{\natexlab{a}})]{hunyuan3d22025tencent}
Tencent~Hunyuan3D Team.
\newblock {Hunyuan3D 2.0}: Scaling diffusion models for high resolution
  textured 3d assets generation.
\newblock \emph{arXiv preprint arXiv:2501.12202}, 2025{\natexlab{a}}.

\bibitem[Team(2025{\natexlab{b}})]{lai2025hunyuan3d25highfidelity3d}
Tencent~Hunyuan3D Team.
\newblock Hunyuan3d 2.5: Towards high-fidelity 3d assets generation with
  ultimate details, 2025{\natexlab{b}}.

\bibitem[Vaswani et~al.(2017)Vaswani, Shazeer, Parmar, Uszkoreit, Jones, Gomez,
  Kaiser, and Polosukhin]{vaswani17attention}
Ashish Vaswani, Noam Shazeer, Niki Parmar, Jakob Uszkoreit, Llion Jones,
  Aidan~N. Gomez, Lukasz Kaiser, and Illia Polosukhin.
\newblock Attention is all you need.
\newblock In \emph{{NeurIPS}}, 2017.

\bibitem[Wang et~al.(2024)Wang, Chen, Huang, Ben, Wang, Mi, Huang, Zhao, Chen,
  Yang, et~al.]{wang2024grutopia}
Hanqing Wang, Jiahe Chen, Wensi Huang, Qingwei Ben, Tai Wang, Boyu Mi, Tao
  Huang, Siheng Zhao, Yilun Chen, Sizhe Yang, et~al.
\newblock {GRUtopia}: Dream general robots in a city at scale.
\newblock \emph{arXiv preprint arXiv:2407.10943}, 2024.

\bibitem[Wang et~al.(2025)Wang, Wang, Hu, Zhang, Yu, and
  Xu]{wang2025kinematify}
Jiawei Wang, Dingyou Wang, Jiaming Hu, Qixuan Zhang, Jingyi Yu, and Lan Xu.
\newblock {Kinematify}: Open-vocabulary synthesis of high-dof articulated
  objects.
\newblock \emph{arXiv preprint arXiv:2511.01294}, 2025.

\bibitem[Wang et~al.(2019)Wang, Sun, Liu, Sarma, Bronstein, and
  Solomon]{wang2019dgcnn}
Yue Wang, Yongbin Sun, Ziwei Liu, Sanjay~E. Sarma, Michael~M. Bronstein, and
  Justin~M. Solomon.
\newblock Dynamic graph cnn for learning on point clouds.
\newblock \emph{ACM TOG}, 2019.

\bibitem[Wei et~al.(2022)Wei, Chabra, Ma, Lassner, Zollhoefer, Rusinkiewicz,
  Sweeney, Newcombe, and Slavcheva]{wei2022nasam}
Fangyin Wei, Rohan Chabra, Lingni Ma, Christoph Lassner, Michael Zollhoefer,
  Szymon Rusinkiewicz, Chris Sweeney, Richard Newcombe, and Mira Slavcheva.
\newblock Self-supervised neural articulated shape and appearance models.
\newblock In \emph{CVPR}, 2022.

\bibitem[Weng et~al.(2024)Weng, Wen, Tremblay, Blukis, Fox, Guibas, and
  Birchfield]{weng2024neural}
Yijia Weng, Bowen Wen, Jonathan Tremblay, Valts Blukis, Dieter Fox, Leonidas
  Guibas, and Stan Birchfield.
\newblock Neural implicit representation for building digital twins of unknown
  articulated objects.
\newblock In \emph{CVPR}, 2024.

\bibitem[Wu et~al.(2025)Wu, Liu, Linli, Huang, Song, Yu, Wu, and
  Lu]{wu2025reartgs}
Di Wu, Liu Liu, Zhou Linli, Anran Huang, Liangtu Song, Qiaojun Yu, Qi Wu, and
  Cewu Lu.
\newblock Reartgs: Reconstructing and generating articulated objects via 3d
  gaussian splatting with geometric and motion constraints.
\newblock In \emph{NeurIPS}, 2025.

\bibitem[Wu et~al.(2023)Wu, Li, Jakab, Rupprecht, and Vedaldi]{wu23magicpony}
Shangzhe Wu, Ruining Li, Tomas Jakab, Christian Rupprecht, and Andrea Vedaldi.
\newblock {MagicPony}: Learning articulated {3D} animals in the wild.
\newblock In \emph{CVPR}, 2023.

\bibitem[Wu et~al.(2022)Wu, Zhong, Tagliasacchi, Cole, and Oztireli]{wu2022d}
Tianhao Wu, Fangcheng Zhong, Andrea Tagliasacchi, Forrester Cole, and Cengiz
  Oztireli.
\newblock D\^{2}nerf: Self-supervised decoupling of dynamic and static objects
  from a monocular video.
\newblock \emph{NeurIPS}, 2022.

\bibitem[Xiang et~al.(2020)Xiang, Qin, Mo, Xia, Zhu, Liu, Liu, Jiang, Yuan,
  Wang, Yi, Chang, Guibas, and Su]{xiang20sapien:}
Fanbo Xiang, Yuzhe Qin, Kaichun Mo, Yikuan Xia, Hao Zhu, Fangchen Liu, Minghua
  Liu, Hanxiao Jiang, Yifu Yuan, He Wang, Li Yi, Angel~X. Chang, Leonidas~J.
  Guibas, and Hao Su.
\newblock {SAPIEN:} {A} simulated part-based interactive environment.
\newblock In \emph{{CVPR}}, 2020.

\bibitem[Xiang et~al.(2025)Xiang, Lv, Xu, Deng, Wang, Zhang, Chen, Tong, and
  Yang]{xiang2024structured}
Jianfeng Xiang, Zelong Lv, Sicheng Xu, Yu Deng, Ruicheng Wang, Bowen Zhang,
  Dong Chen, Xin Tong, and Jiaolong Yang.
\newblock Structured 3d latents for scalable and versatile 3d generation.
\newblock In \emph{CVPR}, 2025.

\bibitem[Xue et~al.(2025)Xue, Chen, Liu, and Sun]{xue2025zerops}
Yuheng Xue, Nenglun Chen, Jun Liu, and Wenyun Sun.
\newblock {ZeroPS}: High-quality cross-modal knowledge transfer for zero-shot
  3d part segmentation.
\newblock In \emph{3DV}, 2025.

\bibitem[Yang et~al.(2024)Yang, Huang, Guo, Lu, Wu, Lam, Cao, and
  Liu]{yang2024sampart3d}
Yunhan Yang, Yukun Huang, Yuan-Chen Guo, Liangjun Lu, Xiaoyang Wu, Edmund~Y
  Lam, Yan-Pei Cao, and Xihui Liu.
\newblock {SAMPart3D}: Segment any part in 3d objects.
\newblock \emph{arXiv preprint arXiv:2411.07184}, 2024.

\bibitem[Zhang et~al.(2023)Zhang, Tang, Niessner, and
  Wonka]{zhang20233dshape2vecset}
Biao Zhang, Jiapeng Tang, Matthias Niessner, and Peter Wonka.
\newblock {3DShape2VecSet}: A 3d shape representation for neural fields and
  generative diffusion models.
\newblock \emph{ACM TOG}, 42\penalty0 (4):\penalty0 1--16, 2023.

\bibitem[Zhou et~al.(2023)Zhou, Gu, Li, Liu, Fang, and Su]{zhou2023partslip++}
Yuchen Zhou, Jiayuan Gu, Xuanlin Li, Minghua Liu, Yunhao Fang, and Hao Su.
\newblock {PartSLIP++}: Enhancing low-shot 3d part segmentation via multi-view
  instance segmentation and maximum likelihood estimation.
\newblock \emph{arXiv preprint arXiv:2312.03015}, 2023.

\end{thebibliography}
